\documentclass[conference]{IEEEtran}
\IEEEoverridecommandlockouts
\usepackage{cite}
\usepackage{amsfonts}
\usepackage{algorithmic}
\usepackage{graphicx}
\usepackage{float}
\usepackage{subfigure}
\usepackage{textcomp}
\usepackage{xcolor}
\usepackage{mathrsfs}
\usepackage{amsmath}
\usepackage{amssymb}  
\usepackage{amsthm}
\usepackage{mathrsfs}
\usepackage{amsfonts}
\usepackage{mathrsfs}
\usepackage[framemethod=tikz]{mdframed}

\def\st{\emph{s.t.\/}}

\def\BibTeX{{\rm B\kern-.05em{\sc i\kern-.025em b}\kern-.08em
    T\kern-.1667em\lower.7ex\hbox{E}\kern-.125emX}}
\begin{document}

\title{Intriguing Frequency Interpretation of Adversarial Robustness for CNNs and ViTs}

\author{
Lu Chen$^\dagger$ \ \ Han Yang$^\ddagger$ \ \ Hu Wang$^\diamond$ \ \ Yuxin Cao$^\ast$ \ \ Shaofeng Li$^\ddagger$ \ \ Yuan Luo$^\dagger$\\
$^\dagger$Shanghai Jiao Tong University, China $^\ddagger$Southeast University, China \\$^\diamond$Mohamed bin Zayed University of Artificial Intelligence, UAE\\$^\ast$National University of Singapore, Singapore 
\\
\texttt{lu.chen@sjtu.edu.cn}, \quad
\texttt{\{yanghan,shaofengli\}@seu.edu.cn},  \\ \texttt{Hu.Wang@mbzuai.ac.ae}, \quad
\texttt{yuxincao@comp.nus.edu.sg
}, \quad 

\texttt{luoyuan@cs.sjtu.edu.cn}
}

\maketitle

\begin{abstract}
Adversarial examples have attracted significant attention over the years, yet understanding their frequency-based characteristics remains insufficient. In this paper, we investigate the intriguing properties of adversarial examples in the frequency domain for the image classification task, with the following key findings.
(1) As the high-frequency components increase, the performance gap between adversarial and natural examples becomes increasingly pronounced. (2) The model performance against filtered adversarial examples initially increases to a peak and declines to its inherent robustness. (3) In Convolutional Neural Networks, mid- and high-frequency components of adversarial examples exhibit their attack capabilities, while in Transformers, low- and mid-frequency components of adversarial examples are particularly effective. These results suggest that different network architectures have different frequency preferences and that differences in frequency components between adversarial and natural examples may directly influence model robustness. Based on our findings, we further conclude with three useful proposals that serve as a valuable reference to the AI model security community.
\end{abstract}

\begin{IEEEkeywords}
adversarial robustness, adversarial perturbations, high-frequency components
\end{IEEEkeywords}

\section{Introduction}
Despite the fact that deep neural networks (DNNs) achieve remarkable performance in many fields~\cite{Devlin19,Vaswani17,Dosovitskiy21}, their counterintuitive vulnerability attracts increasing attention, both for safety-critical applications~\cite{Sharif16,Cao2019AdversarialSA} and the black-box mechanism of DNNs~\cite{Fazlyab19, IIyas19}. DNNs have been found vulnerable to adversarial examples~\cite{Szegedy14,Goodfellow15,moosavi2017unadversarial, Salman2020DoAR}, where small perturbations on the input can easily change the predictions of a well-trained DNN with high confidence.

Since then, how to alleviate the vulnerability of DNNs so as to narrow the performance gap between adversarial/natural examples is another key issue. Existing methods including defensive distillation~\cite{Papernot16} and pixel denoising~\cite{Liao18} have shown their limitations due to follow-up attack strategies~\cite{Nicholas17} or gradient masking~\cite{Athalye19}. Amongst them, adversarial training~\cite{Goodfellow15,Madry18} and its variants~\cite{Zhang19,Wang20} indicate their reliable robustness and outperform~\cite{croce2020reliable}. 
Moreover, as a data augmentation method, adversarial training currently seems to rely on additional data~\cite{Schmidt18,Rebuffi21} to further improve robustness. 

Recalling that high-frequency components can be potentially linked to adversarial examples~\cite{WangHaohan20, Yin19, Harder21, jiang2021focal}, however, few explorations discuss the relationship between frequency components and the attacking capabilities of adversarial examples, \textit{i.e.}, the performance gap between adversarial/natural examples statistically. 
In this paper, we revisit adversarial and natural examples from frequency perspectives. We empirically identify that the performance gap between adversarial and natural examples becomes increasingly pronounced, since higher frequency components of input samples are introduced. Specifically, as high-frequency components of adversarial examples increase, the DNN performance initially rises and subsequently decreases to the adversarial robustness of the DNN. These findings have been validated across various model architectures, including Convolutional Neural Networks (ConvNets) and Vision Transformers (ViTs).\textcolor{black}{ We further confirmed the observations on the CIFAR-10, CIFAR-100~\cite{Krizhevsky09}, and Tiny ImageNet~\cite{le2015tiny} datasets,  as well as across widely adopted attack methods, including the FGSM~\cite{goodfellow2014explaining}, C\&W~\cite{carlini2017towards}, PGD~\cite{Madry18} and AutoAttack~\cite{croce2020reliable} methods.} Furthermore, we conducted a statistical analysis of the frequency differences between adversarial and natural examples in both standard and adversarially-trained models, which shows that the frequency discrepancies are mainly concentrated in the mid-to-high frequency components. These findings suggest that the differences in frequency components between adversarial and natural examples may directly influence model robustness.

\noindent\textbf{Contributions.} In this paper, we reveal intriguing properties of adversarial examples in ConvNets and ViTs from a frequency perspective. (1) We identify that the performance gap between adversarial and natural examples becomes increasingly pronounced as higher frequency components are introduced. (2) The model performance against filtered adversarial examples initially increases to achieve the highest performance, and subsequently decreases to the model robustness. (3) We observe that in ConvNets, the mid- and high-frequency components of adversarial examples reflect their attack capabilities, whereas ViTs exhibit greater sensitivity to the low- and mid-frequency components. (4) We further propose three proposals that offer valuable insights for the AI model security community.

\section{Frequency components of images on DNNs}
\subsection{Preliminaries}
\noindent\textbf{Adversarial attack.} The adversarial attack aims to find a
small perturbation $\delta$ within the $\epsilon$-neighborhood of a natural example $\mathbf{x}$ to maximize the classification loss $\ell$, which misleads the model to misclassify with high confidence.

\begin{equation}\label{eq:adversarialattack}
\delta = \arg\max_{\delta} \ell(f_{\theta}(\mathbf{x}+\mathbf{\delta}), y), \quad \st~||\delta||_p \leq \epsilon,
\end{equation}
\noindent where $f$ denotes a DNN model with parameters $\theta$, and $(\mathbf{x},y)$ denotes a pair of an image $\mathbf{x}$ and its ground-truth label $y$. 

\noindent\textbf{Adversarial training.} 
Adversarial training can be considered as a min-max optimization problem:
\begin{equation}
\theta = \arg\min_{\theta} \mathbb{E}_{\mathbf{x}} [ \max_{||\delta||_p \leq \epsilon}\ell(f_{\theta}(\mathbf{x}+\delta),y) ].
\end{equation} \label{eq:AT}
\noindent The inner maximization problem aims to find the worst-case perturbations to deceive the model, while the outer minimization problem is to optimize model parameters on adversarial examples to improve model robustness.

\noindent\textbf{Discrete Fourier Transform for Images. }
The 2D Discrete Fourier Transform $\mathcal{F}$ converts an two dimensional image $\mathbf{x}\in \mathbb{R}^{M \times N}$ into a complex-valued frequency signal. 
\begin{equation}
\mathcal{F}(u,v) = \sum_{m=0}^{M-1}\sum_{n=0}^{N-1}\mathbf{x}(m,n)e^{-j2\pi(\frac{um}{M} +\frac{vn}{N}) },
\end{equation}
where $\mathcal{F}: \mathbb{R}^{M\times N} \mapsto \mathbb{C}^{M\times N}$ is complex-valued function in the frequency domain. Here, $(m,n)$ represents the coordinate of an image  $\mathbf{x}$ in the spatial domain, and $\mathbf{x}(m,n)$ denotes the pixel value. $(u,v)$ represents the coordinate of the frequency spectrum, and $\mathcal{F}(u, v)$ denotes the complex frequency value. $\mathcal{F}(u, v)$ can be represented as its amplitude $|\mathcal{F}(u, v)|$ and phase $\phi(u, v)$, \textit{i.e.}, $\mathcal{F}(u, v) = |\mathcal{F}(u, v)| e ^{j\phi(u, v)}$. To visualize the amplitude of Fourier spectrum, we shift the low frequency components to the center of the spectrum.

\noindent\textbf{Inverse Discrete Fourier Transform.} 
Given the frequency spectrum $\mathcal{F}$ of an image, the two dimensional image $\mathbf{x}$ can be recovered by applying the inversion of Fourier Transform.
\begin{equation}
\mathbf{x}(m,n) = \frac{1}{MN}\sum_{u=0}^{M-1}\sum_{v=0}^{N-1}\mathcal{F}(u,v)e^{j2\pi(\frac{um}{M} +\frac{vn}{N}) }.
\end{equation}

\subsection{Methodology }


\noindent\textbf{Frequency components of images affecting DNNs.} We aim to investigate whether the differences in frequency components between adversarial and natural examples are associated with their performance gap on models.~\cite{WangHaohan20, Yin19} have indicated that adversarial perturbations may exhibit a higher concentration in the mid- and high-frequency components. Do these differences in mid- and high-frequency components statistically contribute to the performance gap between adversarial/natural examples?

To investigate the relationship between frequency differences and the performance gap, we employ a low-pass filter with varying bandwidth $\mathcal{B}$ to isolate and pass only the low-frequency components of adversarial and natural examples. We focus on how the model performance against adversarial examples evolves as the bandwidth $\mathcal{B}$ increases (\textit{i.e.}, the higher frequency components of the images increase), and examine the trend of the performance gap between the filtered adversarial and natural examples. To further validate the impact of frequency components on model performance, we swap the frequency components between adversarial and natural examples and test accuracy on the the frequency-swapped images. Specifically, for merged adversarial examples, we remain the frequency components of adversarial examples within the bandwidth $\mathcal{B}$, and swap the frequency components outside the bandwidth $\mathcal{B}$ from natural examples. Similarly, for merged natural examples, we swap the frequency components outside the bandwidth $B$ from adversarial examples.



\noindent\textbf{For generating filtered images.}
To validate the impact of different frequency components of images on the model performance, we generate filtered images by applying a low-pass filter with varying bandwidth $\mathcal{B}$. Specifically, we define a low-pass filter $\mathcal{L}_\mathcal{B}=\{0,1\}^{M \times N}$ with bandwidth $\mathcal{B}$ as the operation that only allows low-frequency components  within $\mathcal{B}$ of an image to pass through, while removing high-frequency components outside $\mathcal{B}$, resulting in a blurred image. 
\begin{equation}
\begin{aligned}
\mathcal{L_\mathcal{B}}(u,v)&= \begin{cases}
    
    1,  & r<\frac{\mathcal{B}}{2} \\
    0, & r>\frac{\mathcal{B}}{2} 
\end{cases}, \\
\mathbf{x}^{\text{filtered}}&= \mathcal{F}^{-1}(\mathcal{L}_{\mathcal{B}} \circ \mathcal{F}_{\mathbf{x}}).
\end{aligned}
\end{equation}
\noindent where we allow all frequency components of the frequency spectrum with a radius $r<\frac{\mathcal{B}}{2}$ to pass (\textit{i.e.}, $\mathcal{L}_\mathcal{B}(u,v)=1$), while setting all frequency components outside of the circle with a radius $r>\frac{\mathcal{B}}{2}$ to zero (\textit{i.e.}, $\mathcal{L}_\mathcal{B}(u,v)=0$). Then, we apply low-pass filters with varying bandwidth $\mathcal{B}$ to the frequency spectrum $\mathcal{F}$ of the images $\mathbf{x}$, resulting in the filtered spectrum $\mathcal{L}_\mathcal{B} \circ \mathcal{F}_{\mathbf{x}}$. Finally, the Inverse Discrete Fourier Transform $\mathcal{F}^{-1}$ was performed on the filtered spectrum $\mathcal{L}_\mathcal{B} \circ \mathcal{F}_{\mathbf{x}}$ to obtain filtered images $\mathbf{x}^{\text{filtered}}$. We set $\mathcal{B}/M$ as the frequency scale in experiments, where $\mathcal{B}/M=1.0$ denotes the passed frequency range is tangent to the image.

\noindent\textbf{For generating adversarial examples.} 
We generate the adversarial examples using the widely adopted \textcolor{black}{methods, including the Fast Gradient Sign Method (FGSM) ~\cite{goodfellow2014explaining}, C\&W ~\cite{carlini2017towards}, Projected Gradient Descent (PGD)~\cite{Madry18} and AutoAttack~\cite{croce2020reliable} methods,} to construct adversarial perturbations for each image in the test set. Specifically, 
we evaluate the robustness \textcolor{black}{of ConvNets and ViTs}  against filtered adversarial examples on the \textcolor{black}{CIFAR-10, CIFAR-100~\cite{Krizhevsky09}, and Tiny ImageNet~\cite{le2015tiny} datasets.} \textcolor{black}{For $\ell_\infty$-bounded adversarial examples, we use FGSM, PGD-${20}$ (step size $1/255$), and AutoAttack with a maximum perturbation of $\epsilon = 8/255$. Additionally, $\ell_2$-bounded adversarial examples are generated using the C\&W method with parameters $c=100$ and $\kappa=0$~\cite{carlini2017towards}.}


To further evaluate the robustness on the adversarially-trained models~\cite{Madry18}, we train models using the $\ell_{\infty}$ bounded adversarial examples generated by PGD-${10}$ (step size $2/255$).

\section{Experiments}
\label{gen_inst}
\begin{figure*}[htp]
    \centering

    \subfigure[ResNet-18]{\includegraphics[width=0.193\linewidth]{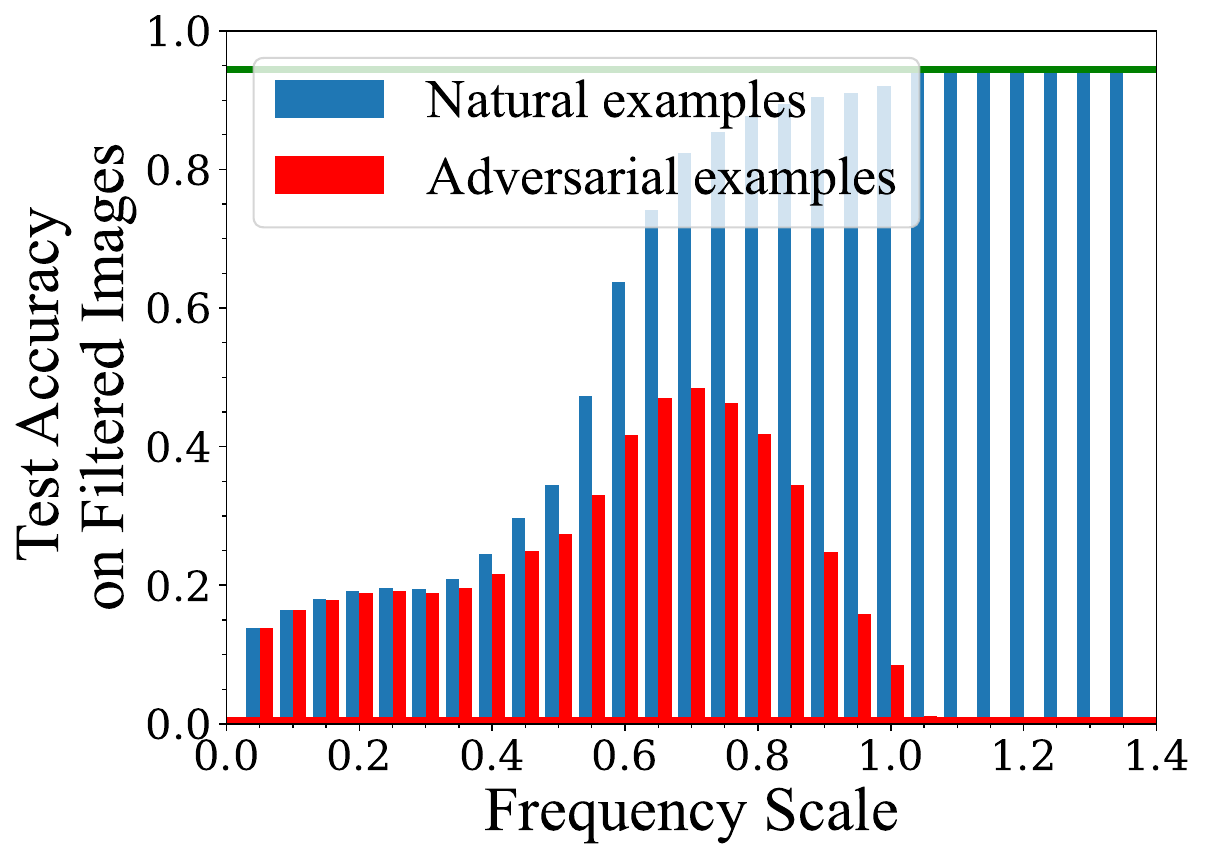}\label{fig:hfca}}
    \subfigure[Wide ResNet-28-10]{\includegraphics[width=0.193\linewidth]{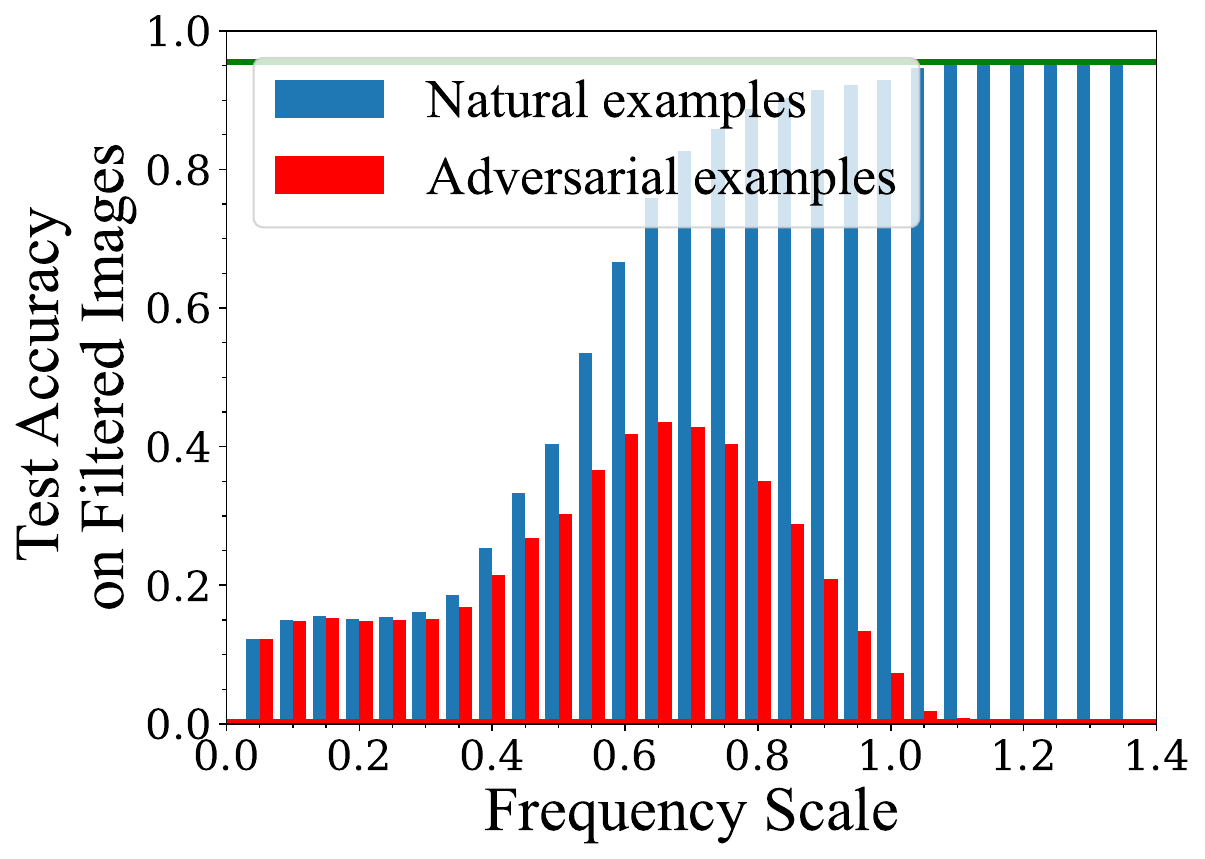}}\label{fig:hfcb} 
    \subfigure[VGG-16]{\includegraphics[width=0.193\linewidth]{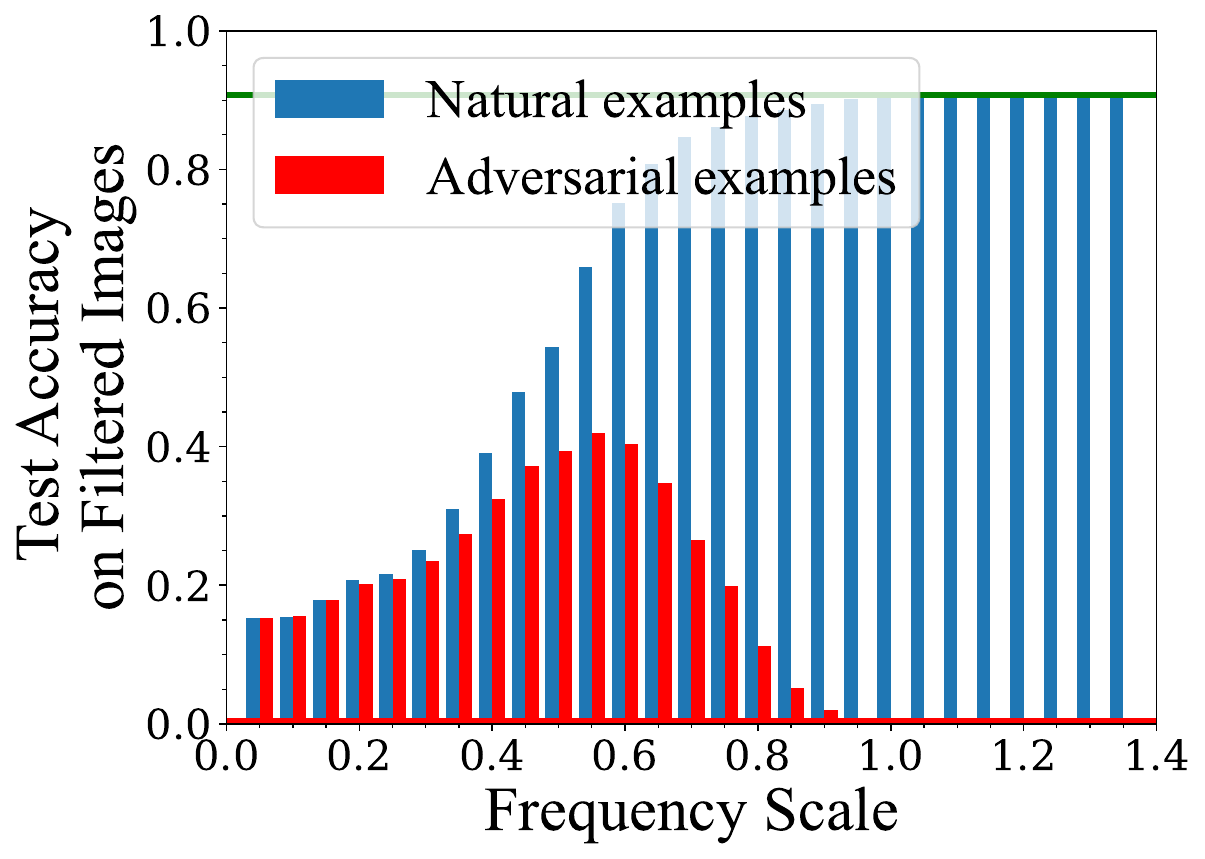}} 
    \subfigure[DenseNet]{\includegraphics[width=0.193\linewidth]{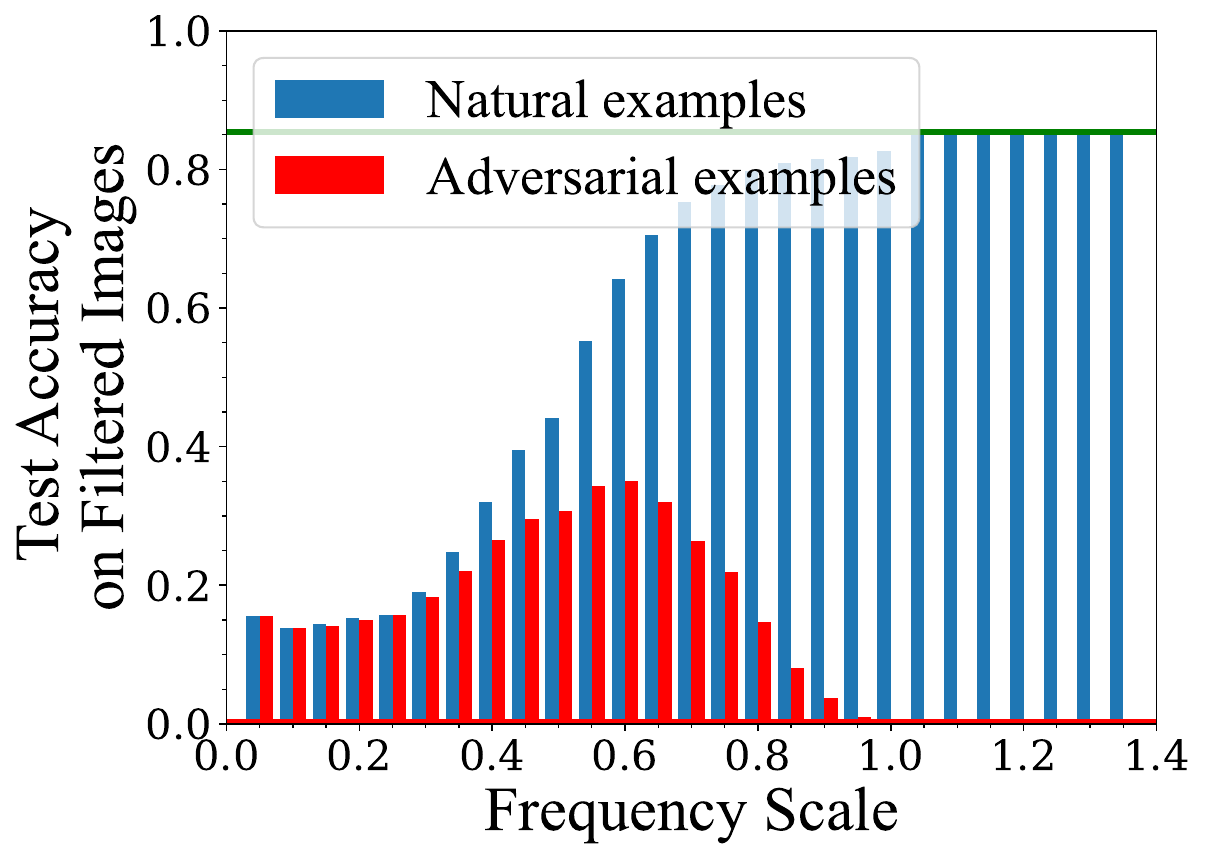}} 
    \subfigure[Performance Gap]{\includegraphics[width=0.193\linewidth]{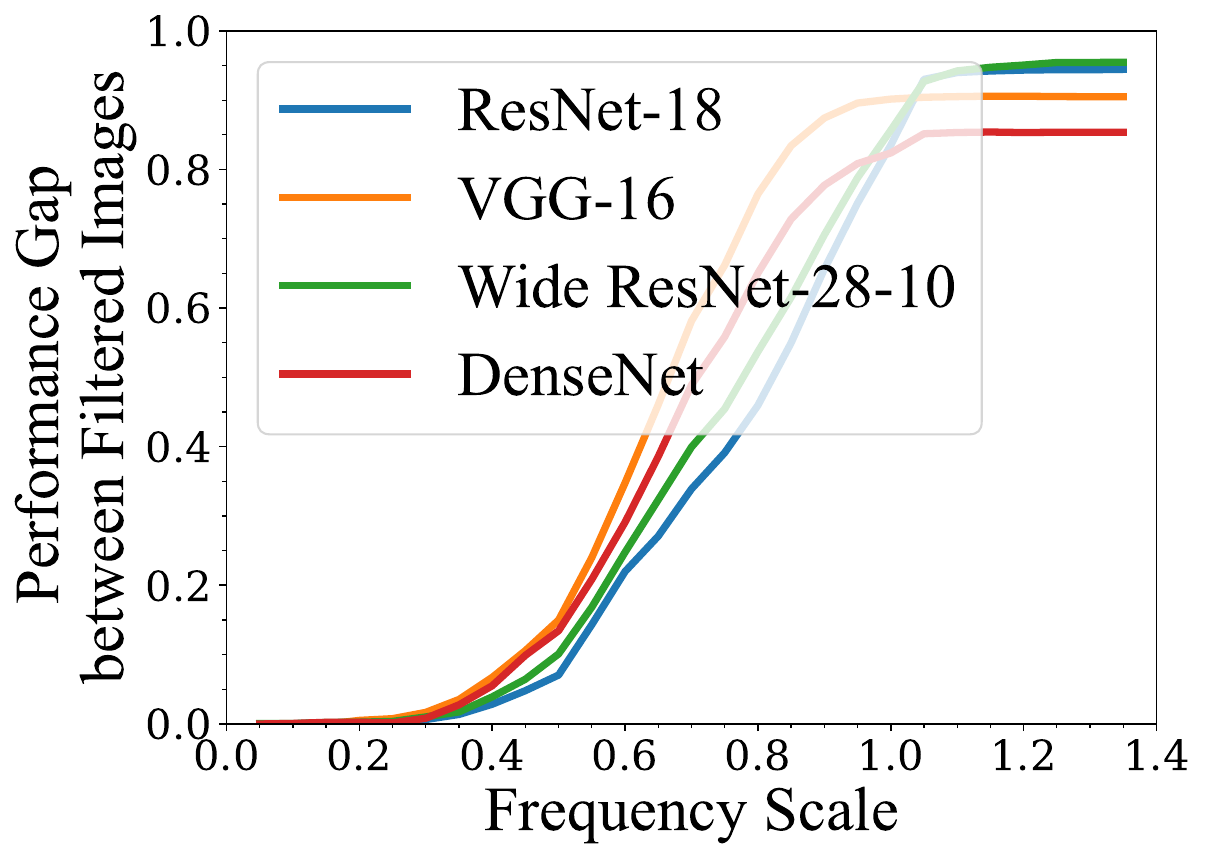}} 
    \vspace{-0.2cm}
    \caption{The differences of frequency components contribute to the performance gap between adversarial and natural examples on ConvNets\textcolor{black}{, as evaluated on the CIFAR-10 dataset}. We tested the accuracy of the filtered adversarial/natural images on (a)-(d) different ConvNets, by employing a low-pass filter with varying bandwidth on the frequency spectrum of images. (e) As higher frequency components are introduced, the performance gap between adversarial and natural examples increases.}
    \label{fig:ConvNets}
    \vspace{-0.1cm}
\end{figure*}

\begin{figure*}[htp]
    \centering

    \subfigure[CIFAR-100 (ResNet-50)]{\includegraphics[width=0.193\linewidth]{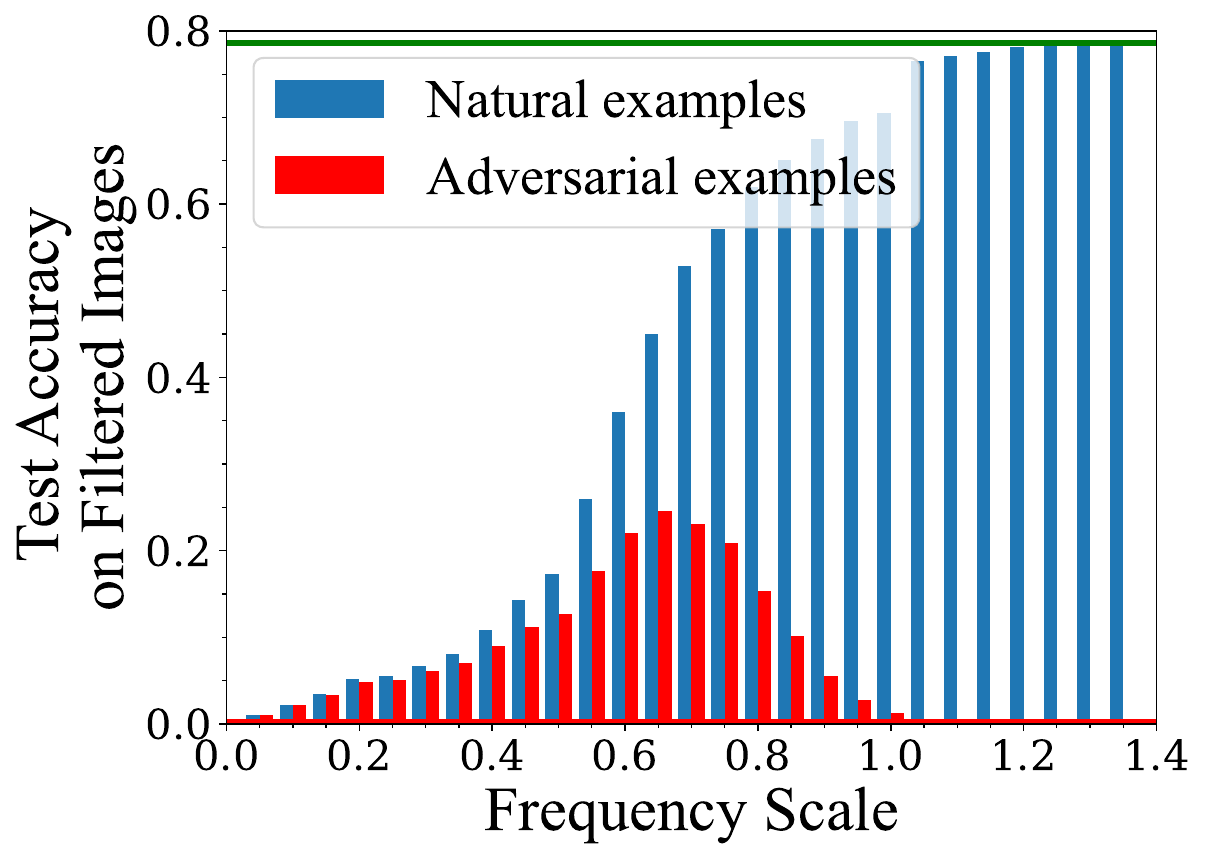}\label{fig:hfca}}
    \subfigure[CIFAR-100 (WRN-28-10)]{\includegraphics[width=0.193\linewidth]{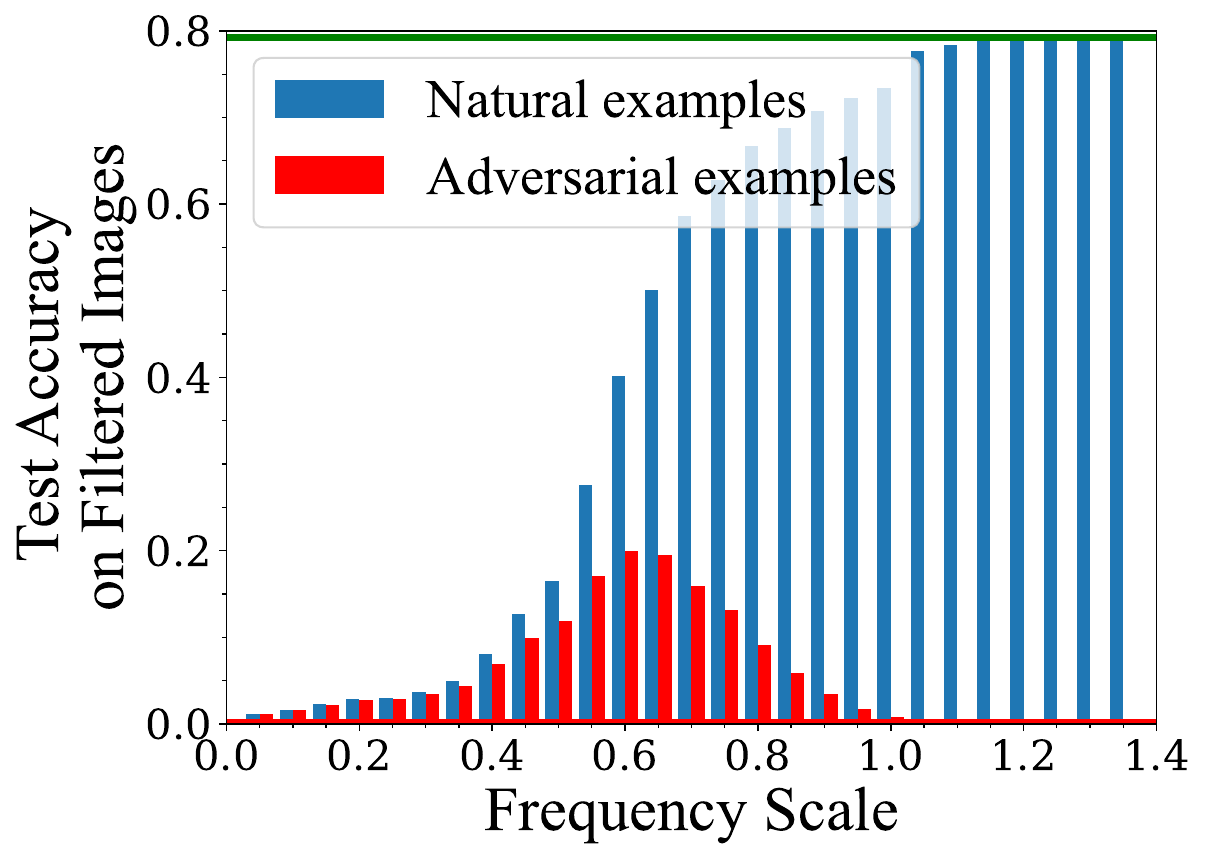}}\label{fig:hfcb} 
    \subfigure[Tiny ImageNet (ResNet-50)]{\includegraphics[width=0.193\linewidth]{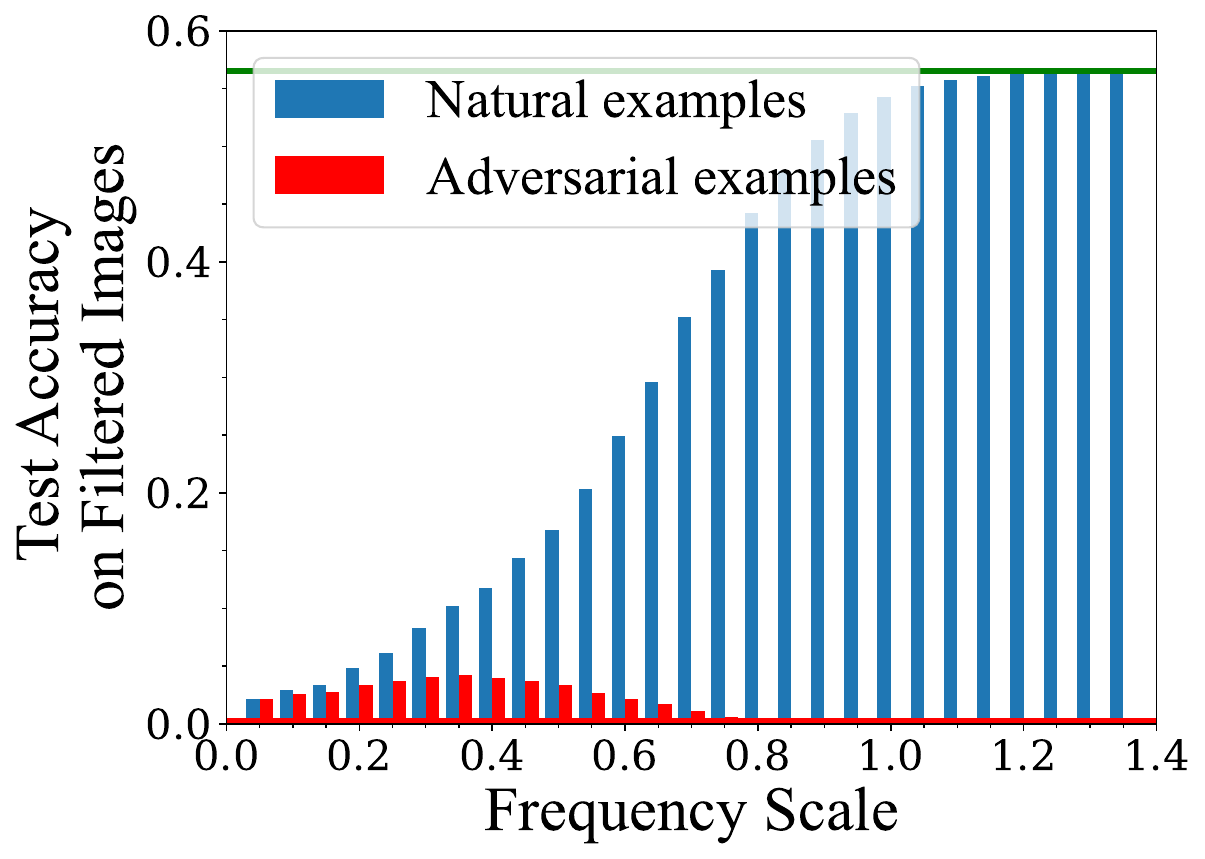}} 
    \subfigure[Tiny ImageNet (WRN-28-10)]{\includegraphics[width=0.193\linewidth]{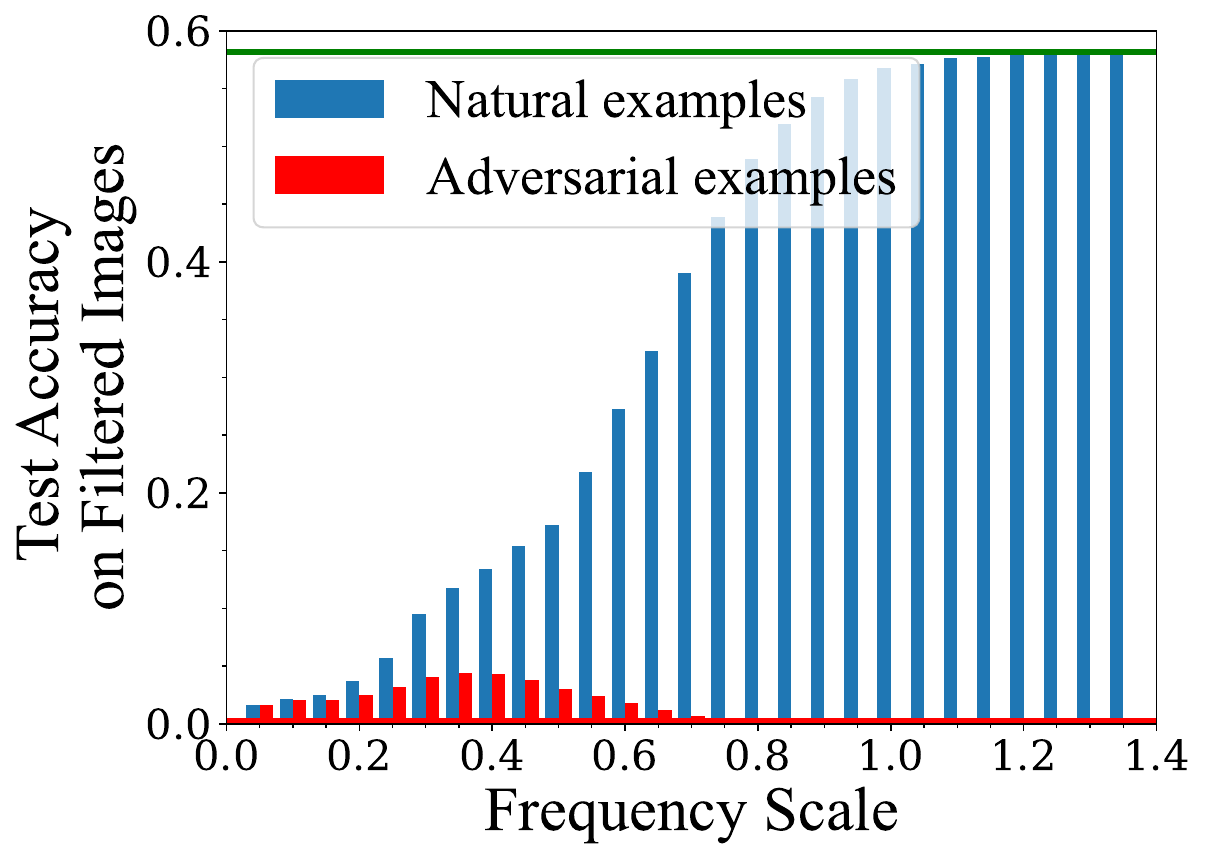}} 
    \subfigure[Performance Gap]{\includegraphics[width=0.193\linewidth]{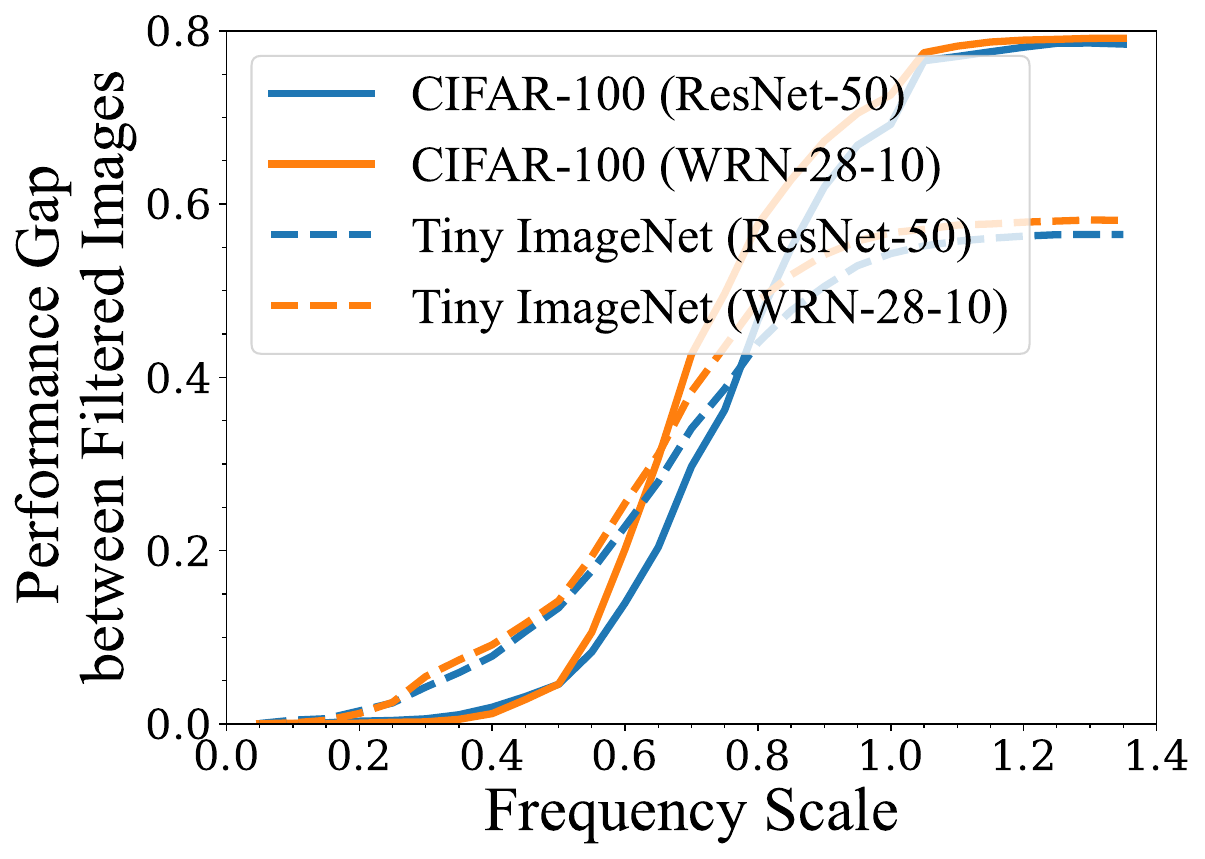}} 
    \vspace{-0.2cm}
    \caption{The differences of frequency components contribute to the performance gap between adversarial and natural examples on ConvNets \textcolor{black}{across multiple datasets}. We evaluated the accuracy of the filtered adversarial/natural images on \textcolor{black}{(a)-(b) the CIFAR-100 dateset and (c)-(d) the Tiny ImageNet dataset}. (e) As higher frequency components are introduced, the performance gap between adversarial and natural examples increases \textcolor{black}{across these datasets}.}
    \label{fig:ConvNets_datasets}
    \vspace{-0.1cm}
\end{figure*}

\begin{figure*}[htp]
    \centering
\vspace{-0.2cm}
    \subfigure[Attack methods on CIFAR-10]{\includegraphics[width=0.24\linewidth]{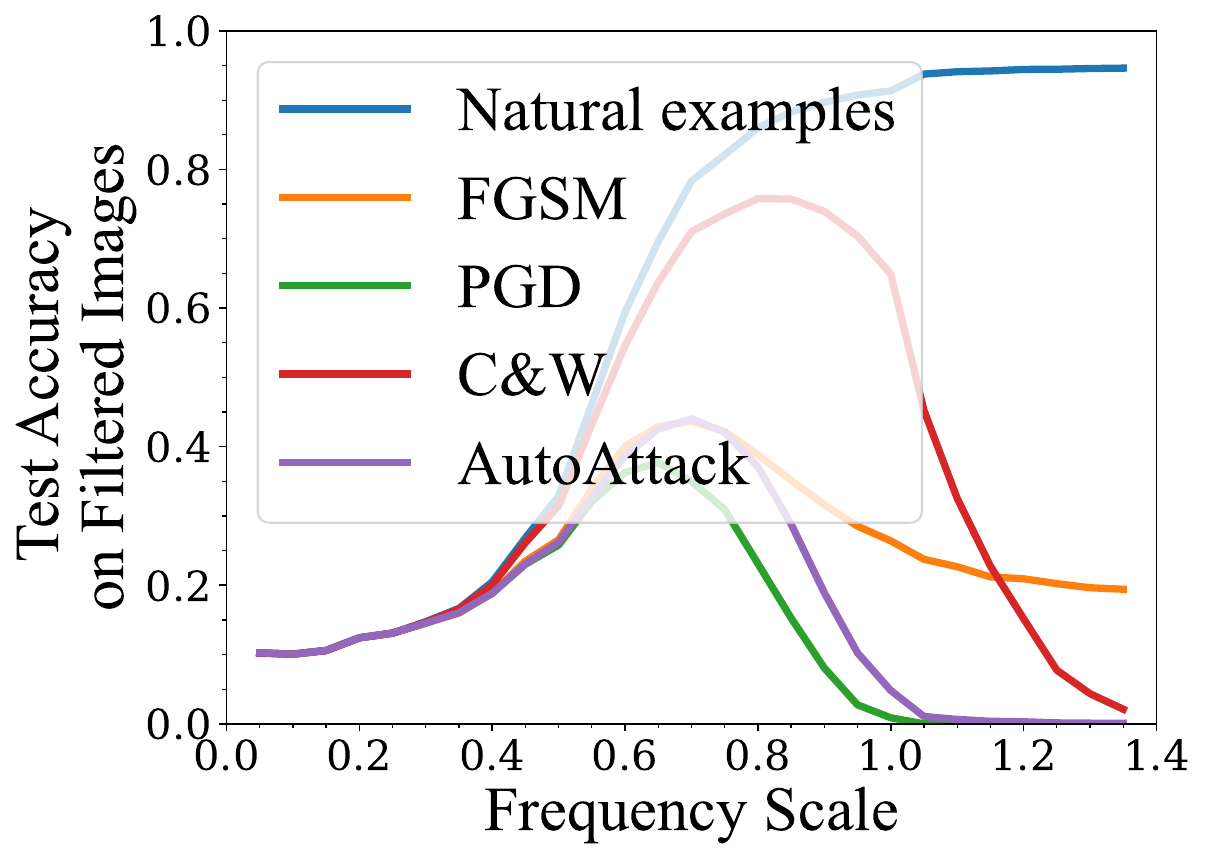}\label{fig:hfca}}
    \subfigure[Attack methods on CIFAR-100]{\includegraphics[width=0.24\linewidth]{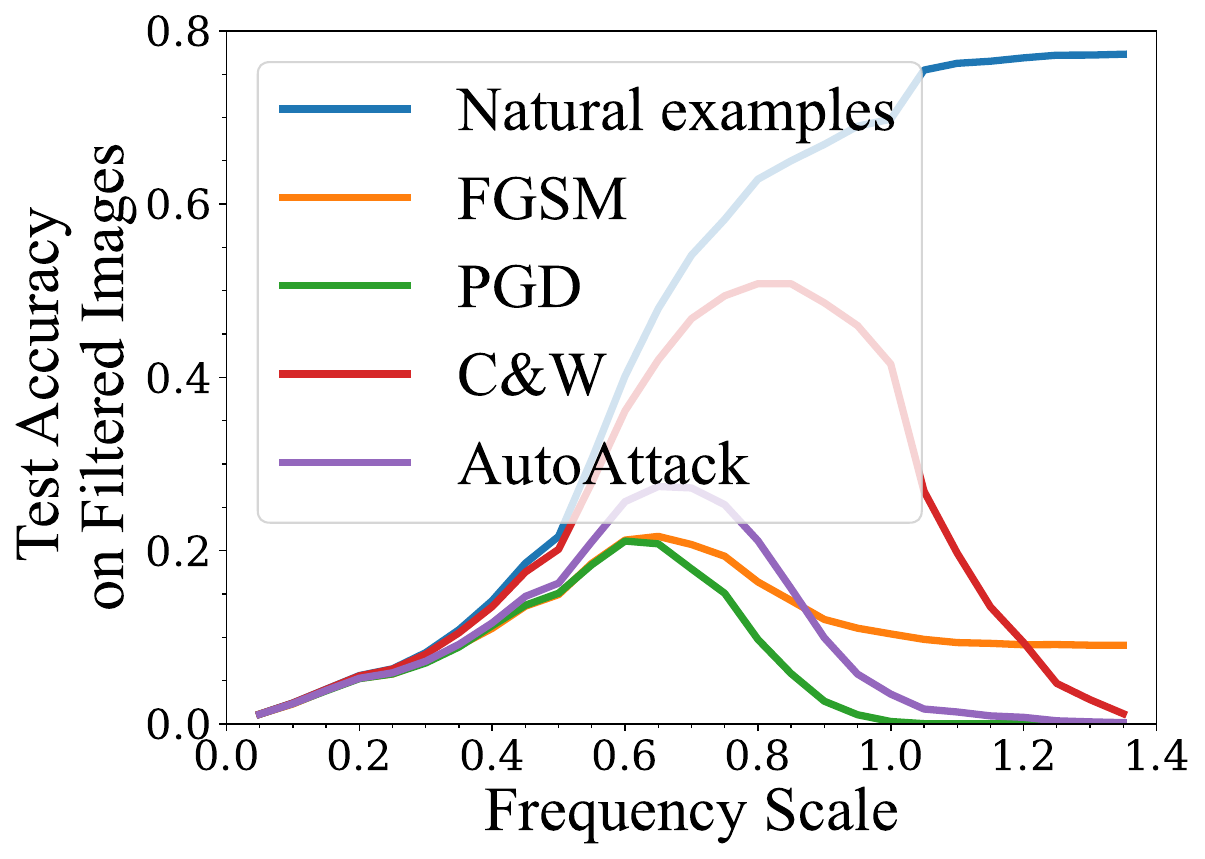}} 
    \subfigure[Attack methods on Tiny ImageNet]{\includegraphics[width=0.24\linewidth]{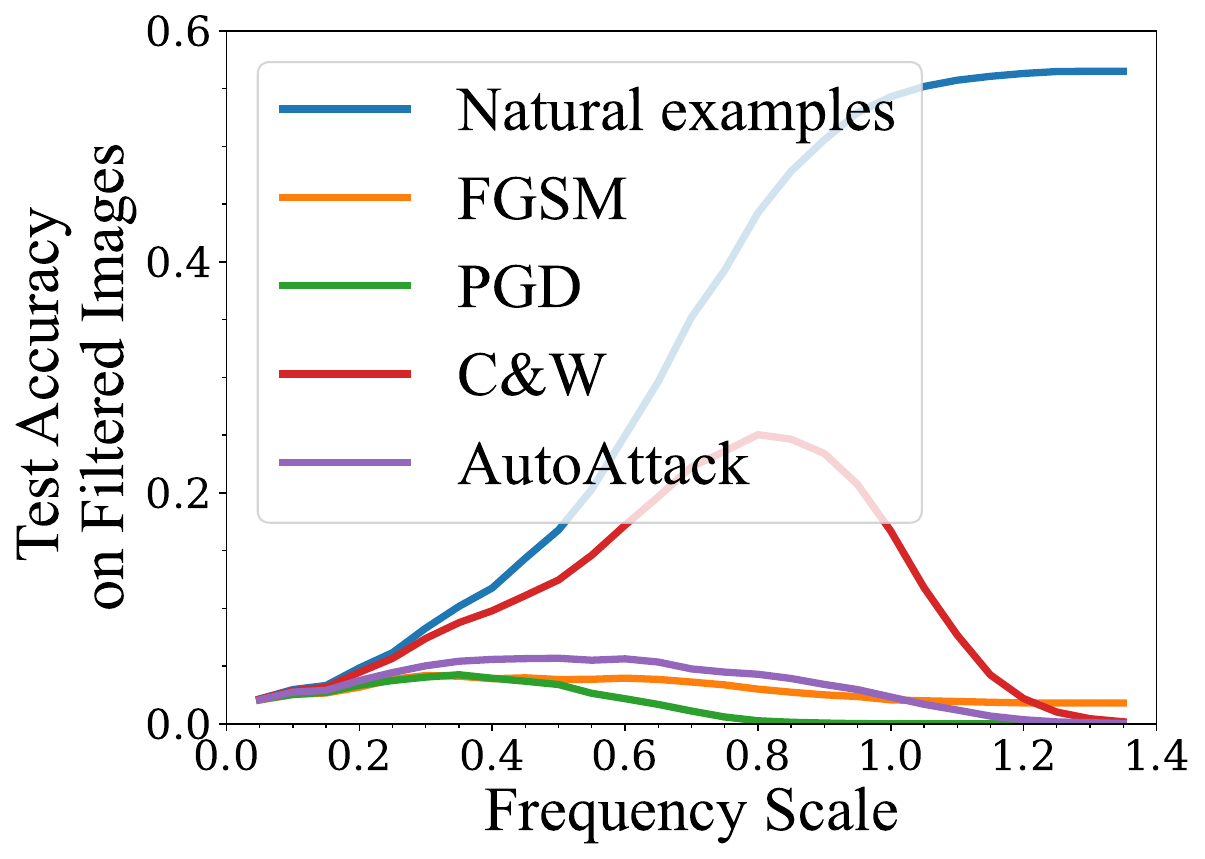}}\label{fig:hfcb} 
    \subfigure[Performance Gap]{\includegraphics[width=0.24\linewidth]{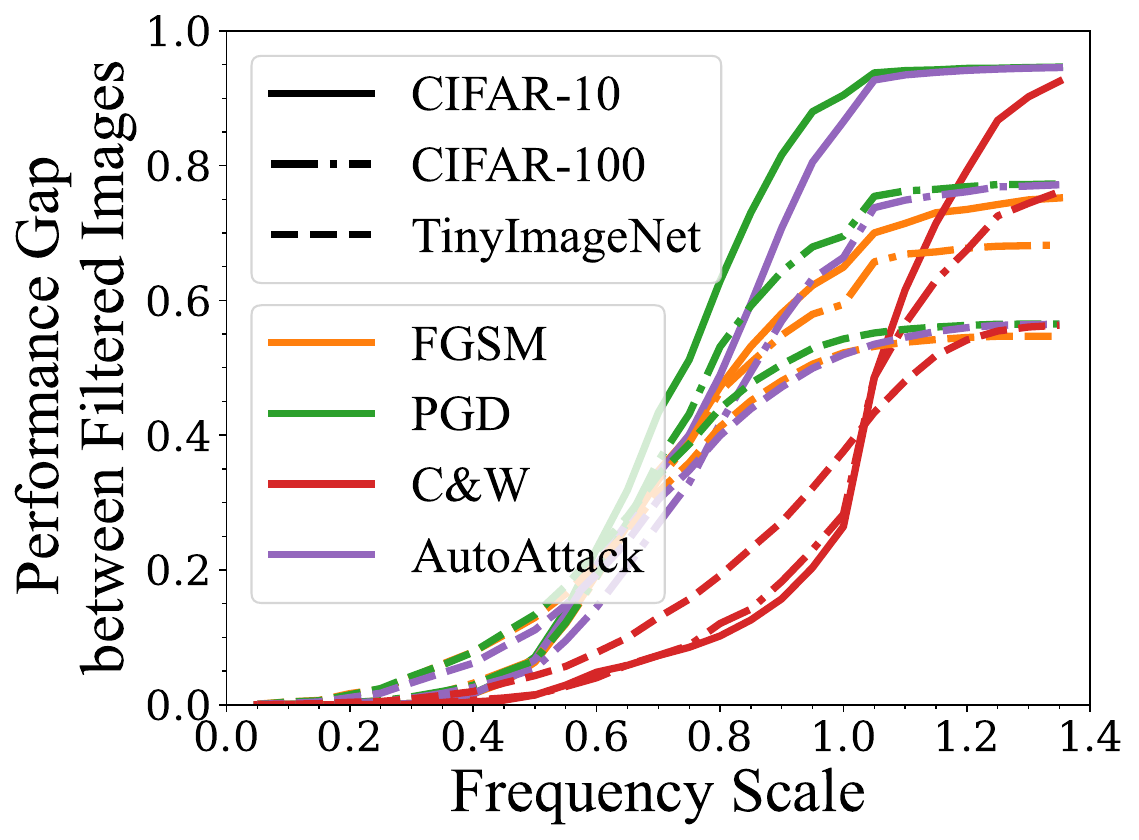}} 
    \vspace{-0.2cm}
    \caption{The differences of frequency components contribute to the performance gap between adversarial and natural examples  on ConvNets \textcolor{black}{across various attack methods}. We evaluated the accuracy of the filtered adversarial examples \textcolor{black}{generated using four different attack methods, the FGSM, C\&W, PGD, and AutoAttack methods, on multiple datasets}. (d) As higher frequency components are introduced, the performance gap between adversarial and natural examples increases \textcolor{black}{across these attack methods}.}
    \vspace{-0.2cm}
    \label{fig:ConvNets_attacks}
\end{figure*}

In this section, we investigate whether the differences in frequency components between adversarial and natural examples contribute to their performance gap on models. 

\subsection{Research Questions on  Frequency Components of Images}

We aim to investigate whether the performance of DNNs is really affected by the frequency components of adversarial and natural examples. Specifically, this section will address the following three questions:

\textbf{\textit{Q1}}: How do the frequency components of adversarial examples affect the performance of convolutional neural networks? 

\textbf{\textit{Q2}}: What impact do the frequency components of adversarial examples have on the effectiveness of Vision Transformers? 

\textbf{\textit{Q3}}: To what extent do the frequency components of adversarial examples shape the performance of adversarially-trained models?


\subsection{Frequency Components of images on ConvNets for \textbf{Q1}}

We evaluate the model performance \textit{w.r.t.} the filtered adversarial/natural examples using various ConvNets, including ResNet-18~\cite{He16}, Wide ResNet-28-10~\cite{zagoruyko2016wide}, VGG-16~\cite{simonyan2014very}, and DenseNet~\cite{huang2017densely}, on the CIFAR-10 dataset~\cite{Krizhevsky09}. 
Fig.~\ref{fig:ConvNets} shows that the model performance \textit{w.r.t.} the filtered adversarial examples \textcolor{black}{generated using the PGD method, as well as natural examples,} on different ConvNets. For standard ConvNets, as higher frequency components are introduced, the performance of filtered adversarial examples initially increases, peaking at an accuracy around 40\%, and subsequently declines to 0.0\% (red line) which reflects the DNN’s robustness against adversarial examples. In contrast, the introduction of higher frequency components in natural examples enhances classification performance, eventually reaching the clean performance of the model (green line). Fig.~\ref{fig:ConvNets}(e) shows the performance gap between filtered adversarial examples and filtered natural examples on various ConvNets. It shows that the low-frequency components of both adversarial and natural examples have a nearly identical impact on the model. However, as higher frequency components are incorporated, the performance gap between filtered adversarial and natural examples gradually increases from near zero to the gap between model generalization and robustness. 

\begin{figure*}[htp]
    \centering
\vspace{-0.2cm}
    \subfigure[ViT-B/16]{\includegraphics[width=0.24\linewidth]{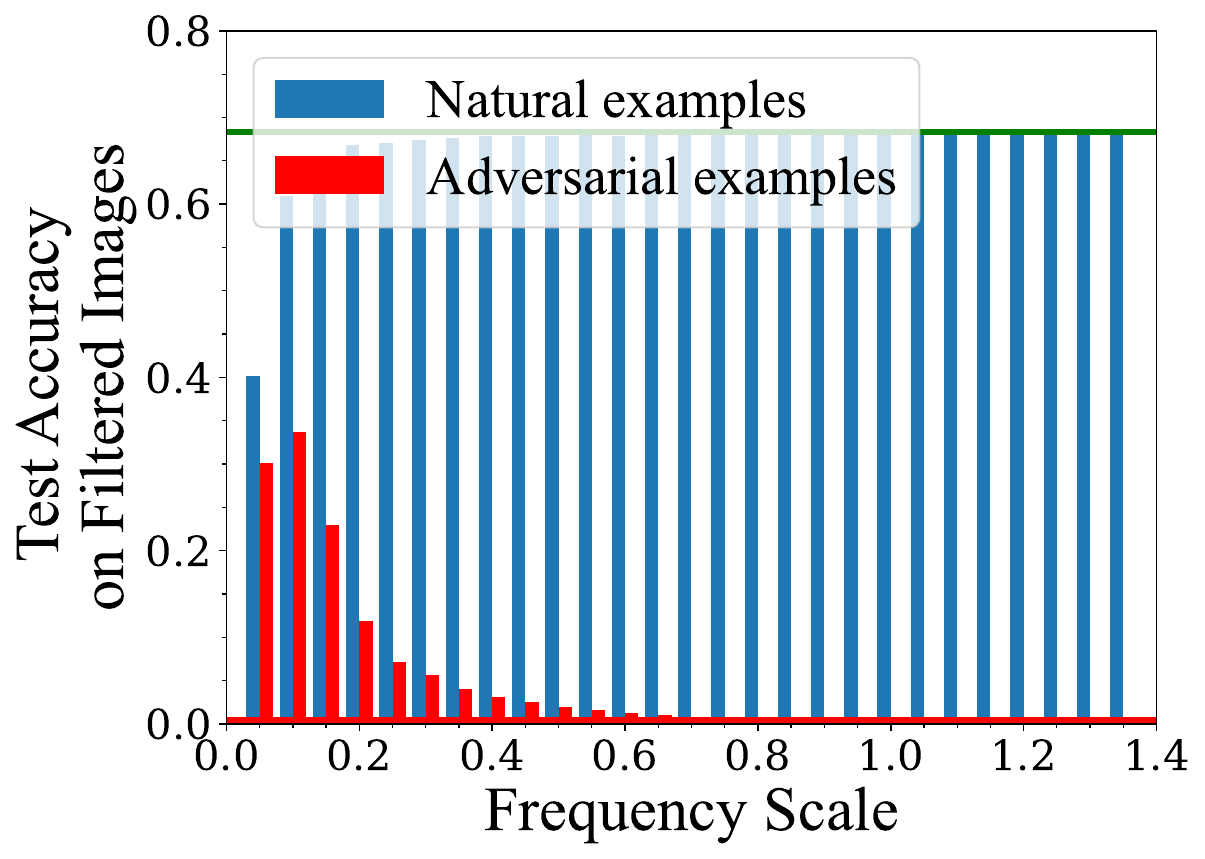}\label{fig:hfca}}
    \subfigure[Swin-B]{\includegraphics[width=0.24\linewidth]{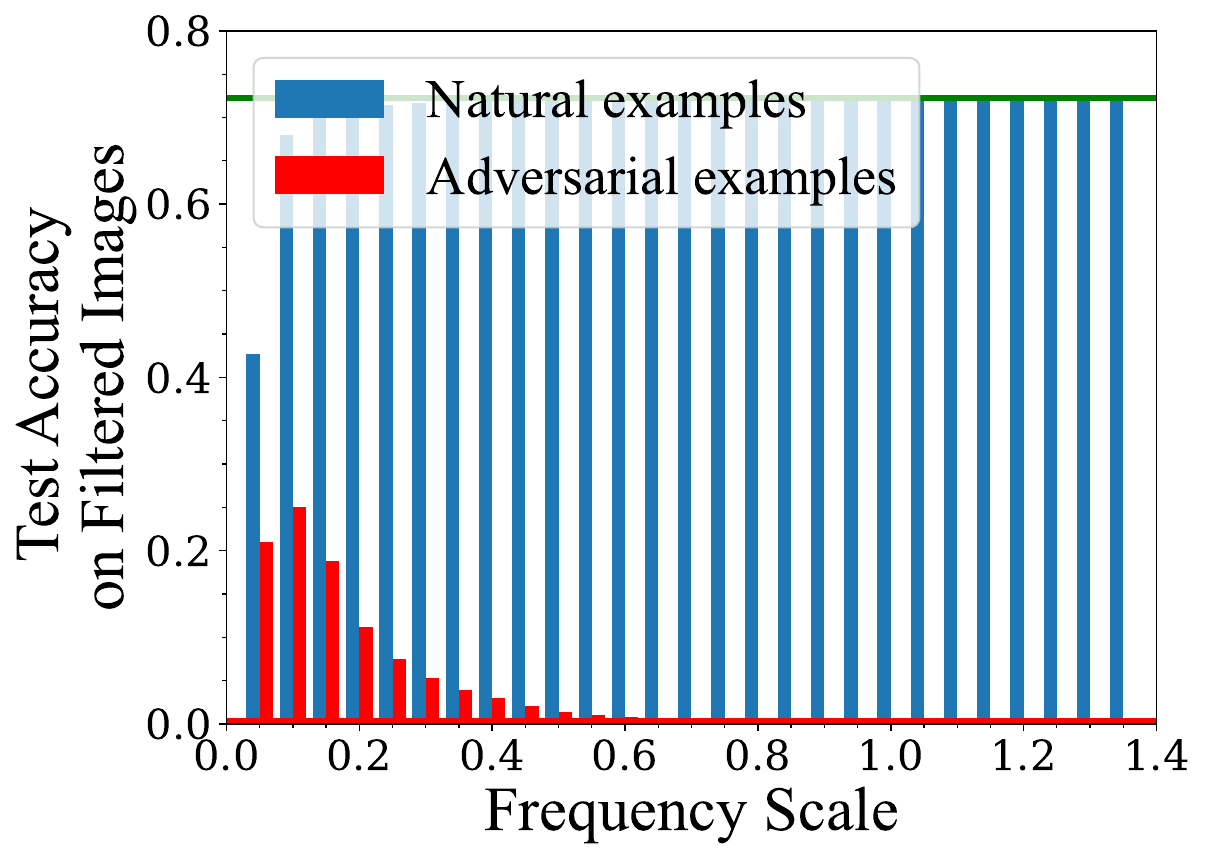}} 
    \subfigure[DeiT-B]{\includegraphics[width=0.24\linewidth]{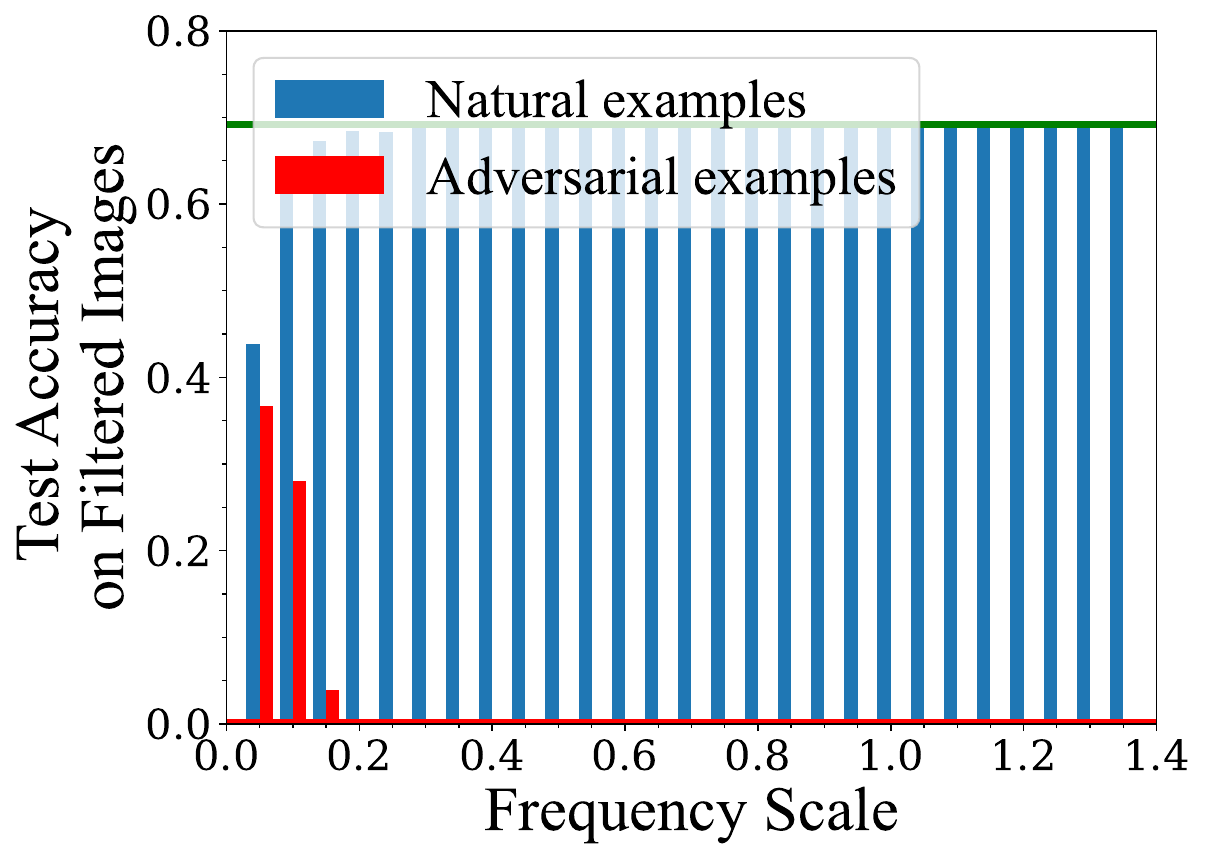}}\label{fig:hfcb} 
    \subfigure[Performance Gap]{\includegraphics[width=0.24\linewidth]{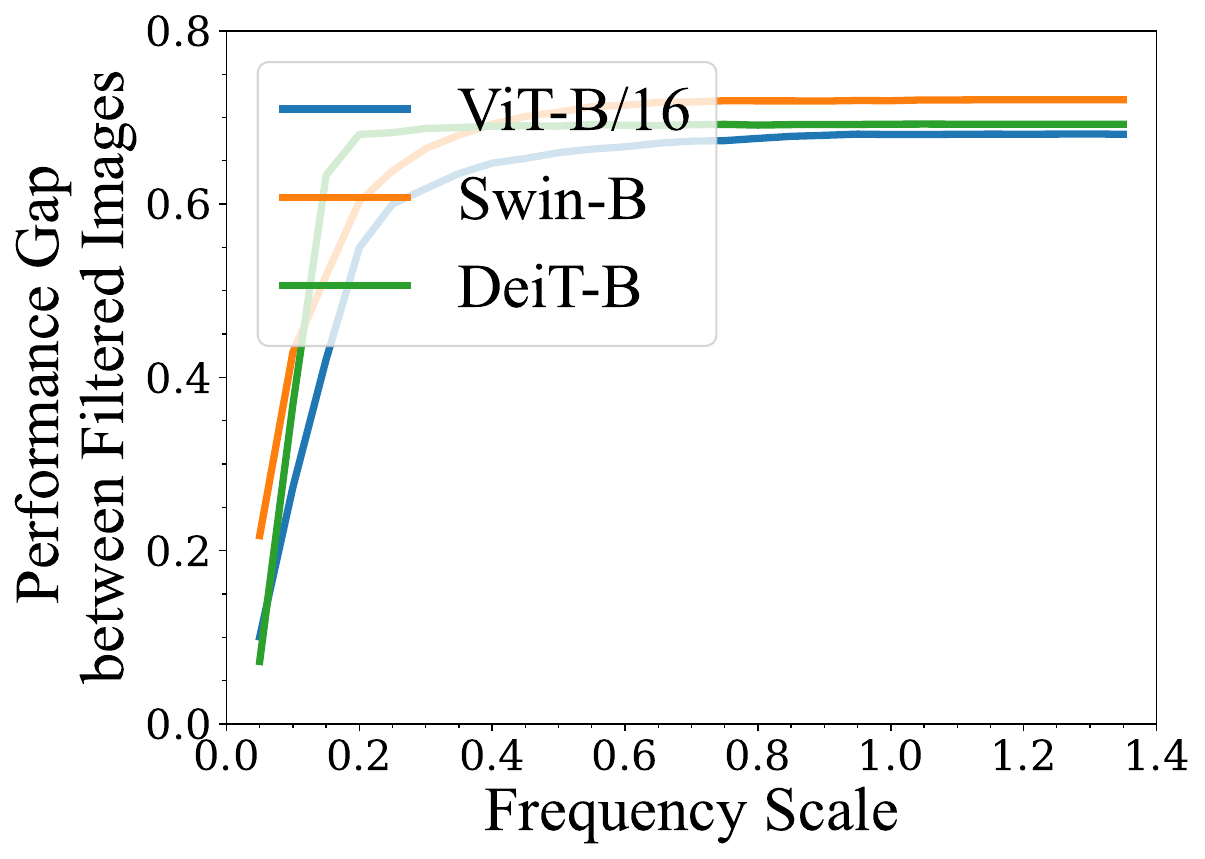}} 
    \vspace{-0.2cm}
    \caption{The differences of frequency components contribute to the performance gap between adversarial and natural examples on ViTs\textcolor{black}{, as evaluated on the CIFAR-10 dataset}. We tested the accuracy of the filtered adversarial/natural images on (a)-(c) different Transformers, by employing a low-pass filter with varying bandwidth on the frequency spectrum of images. (d) As higher frequency components are introduced, the performance gap between adversarial and natural examples increases.}
    \vspace{-0.2cm}
    \label{fig:Transformers}
\end{figure*}

\begin{figure*}[htp]
    \centering

    \subfigure[CIFAR-100 (ViT-B/16)]{\includegraphics[width=0.193\linewidth]{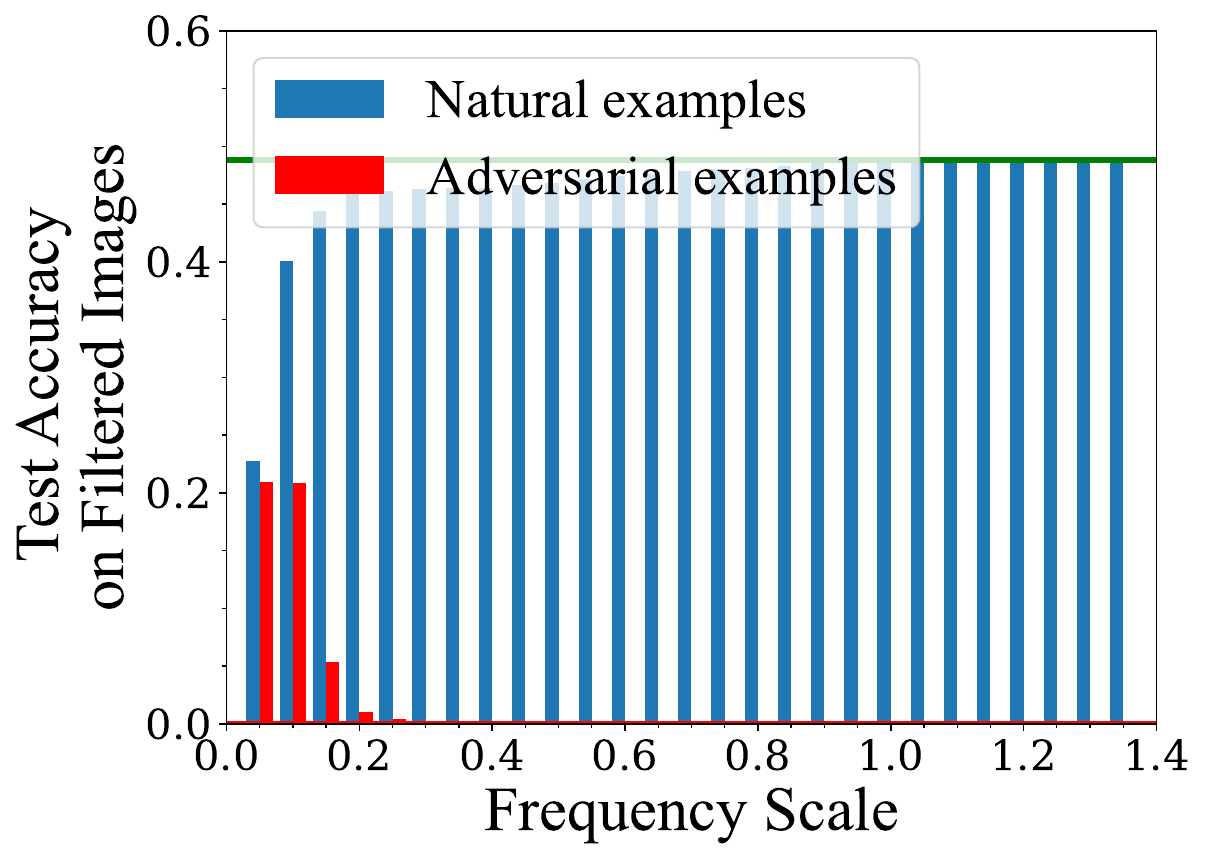}\label{fig:hfca}}
    \subfigure[CIFAR-100 (Swin-B)]{\includegraphics[width=0.193\linewidth]{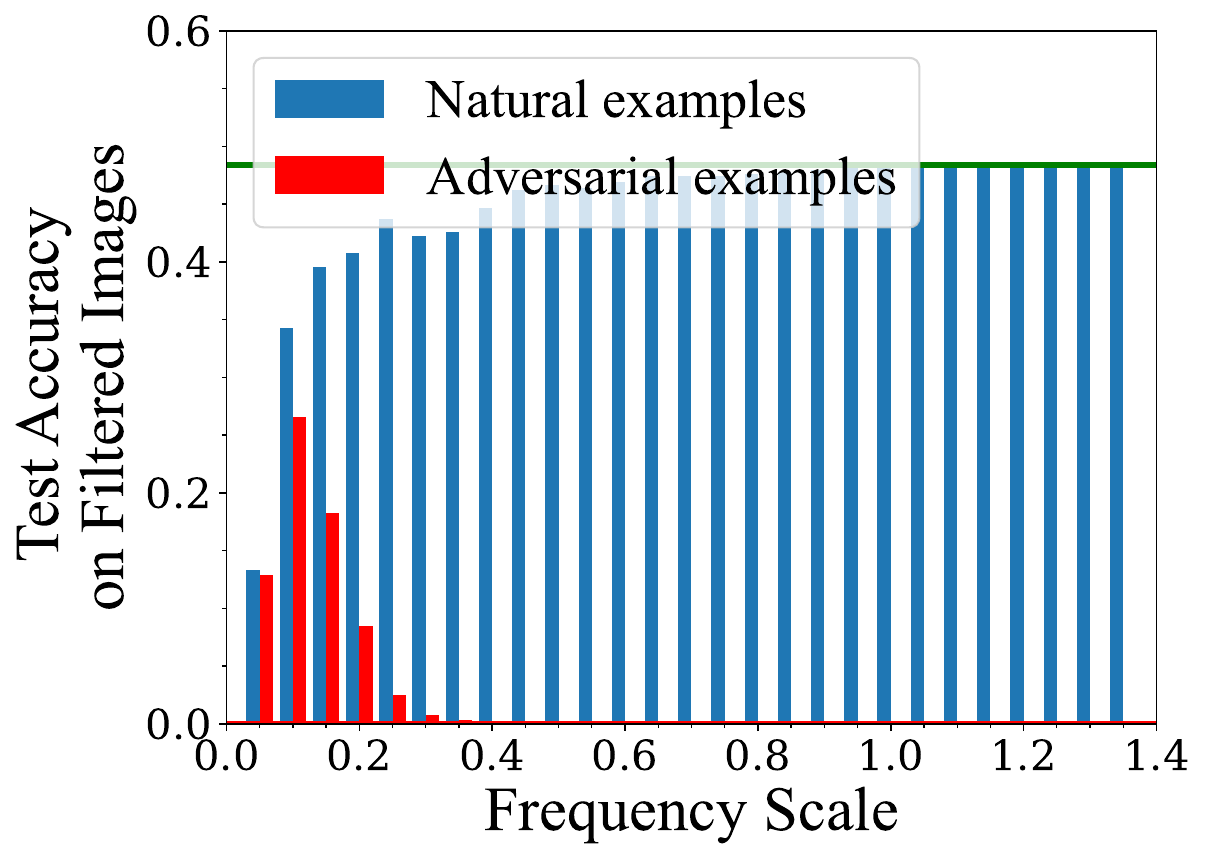}}\label{fig:hfcb} 
    \subfigure[Tiny ImageNet (ViT-B/16)]{\includegraphics[width=0.193\linewidth]{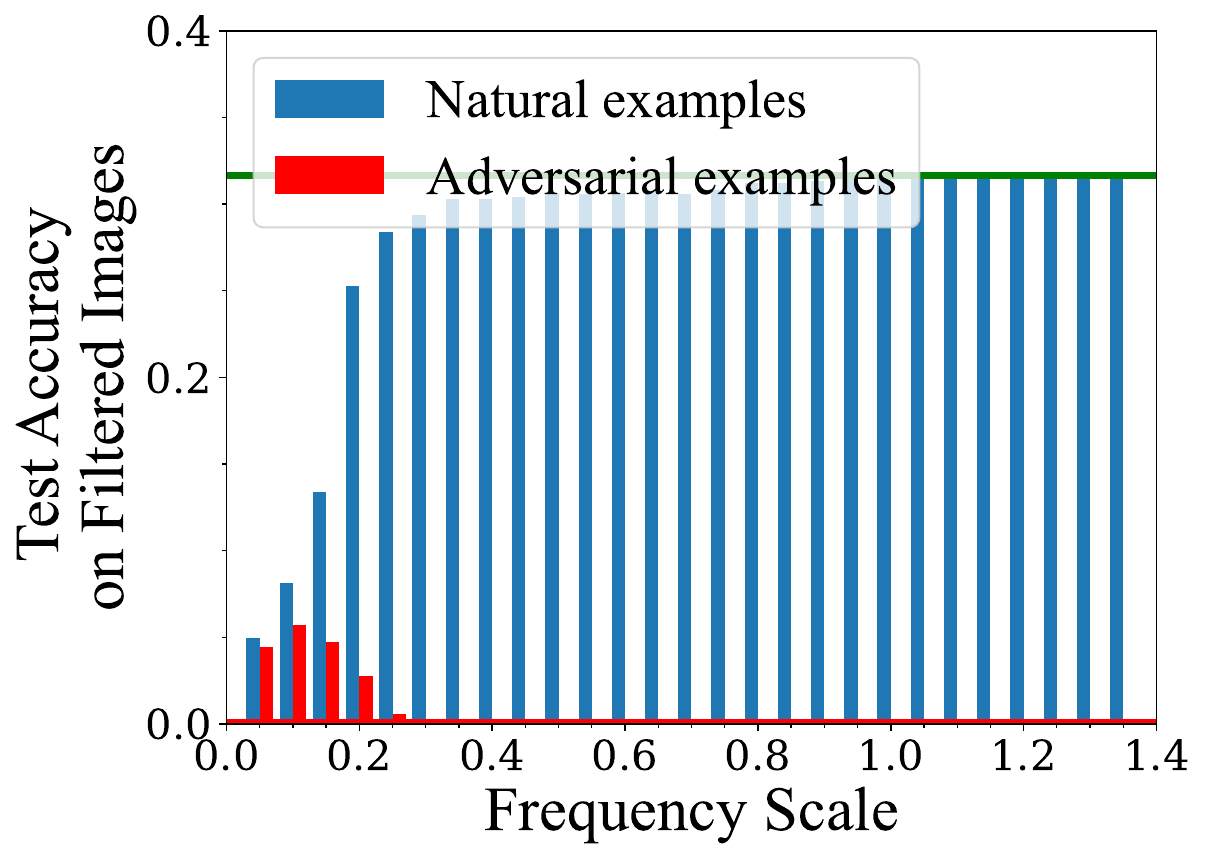}} 
    \subfigure[Tiny ImageNet (Swin-B)]{\includegraphics[width=0.193\linewidth]{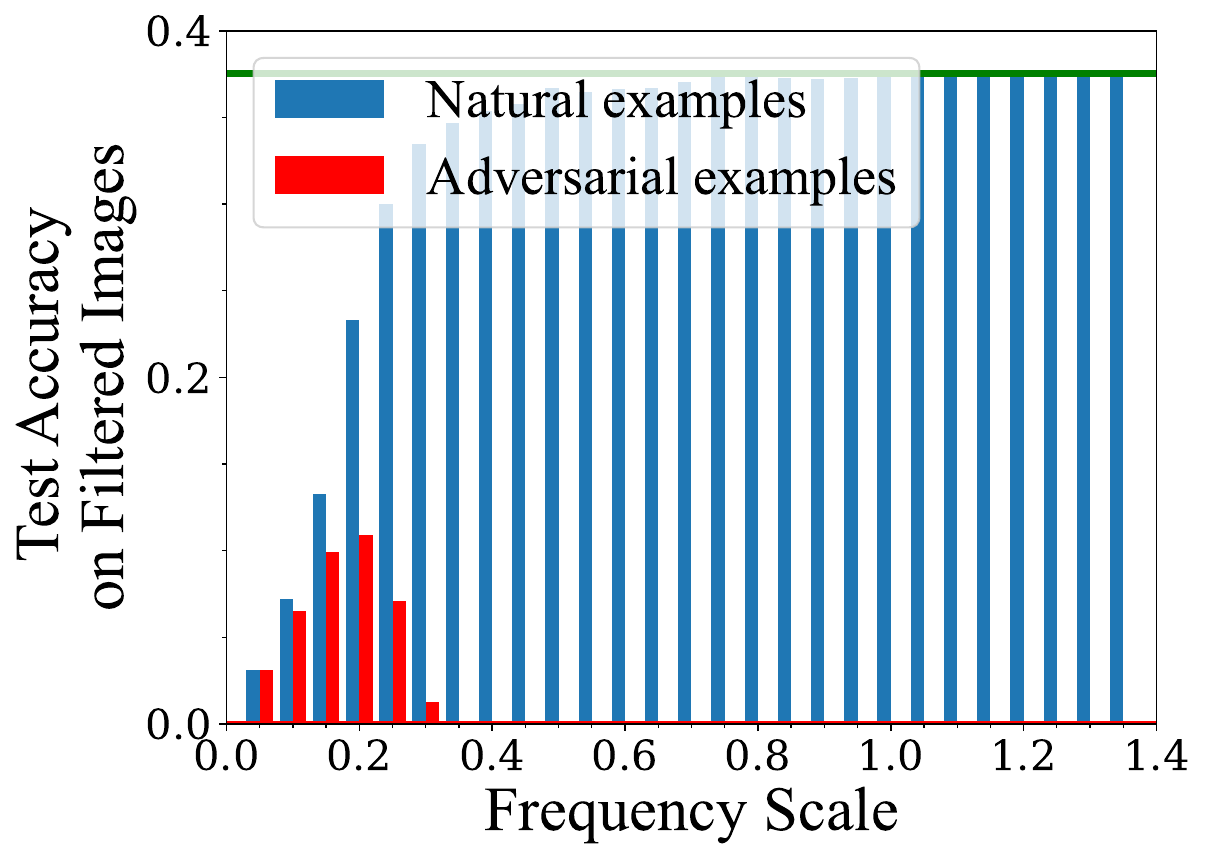}}
    \subfigure[Performance Gap]{\includegraphics[width=0.193\linewidth]{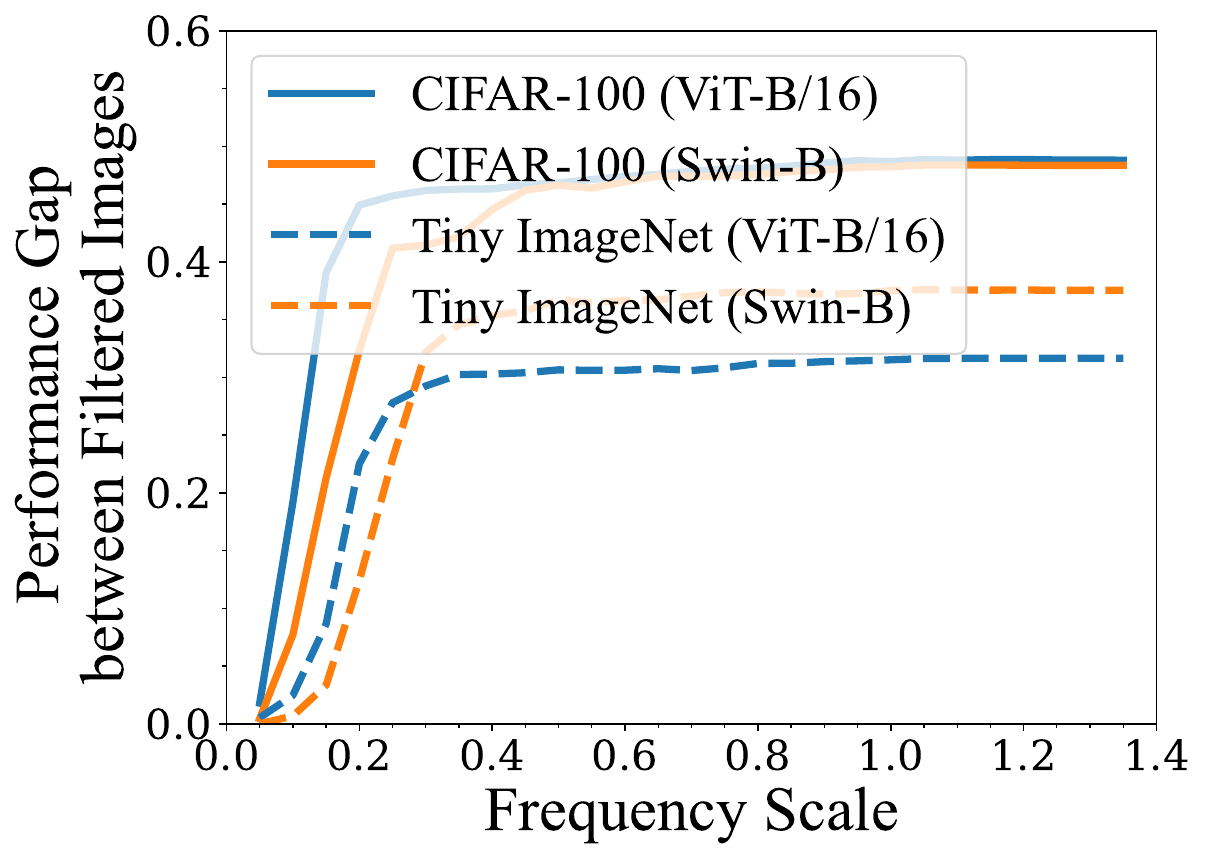}} 
    \vspace{-0.2cm}
    \caption{The differences of frequency components contribute to the performance gap between adversarial and natural examples on ViTs \textcolor{black}{across multiple datasets}. We evaluated the accuracy of the filtered adversarial/natural images on \textcolor{black}{(a)-(b) the CIFAR-100 dateset and (c)-(d) the Tiny ImageNet dataset}. (e) As higher frequency components are introduced, the performance gap between adversarial and natural examples increases \textcolor{black}{across these datasets}.}
    \label{fig:ViTs_datasets}
    \vspace{-0.1cm}
\end{figure*}

\begin{figure*}[htp]
    \centering
\vspace{-0.2cm}
    \subfigure[Attack methods on CIFAR-10]{\includegraphics[width=0.24\linewidth]{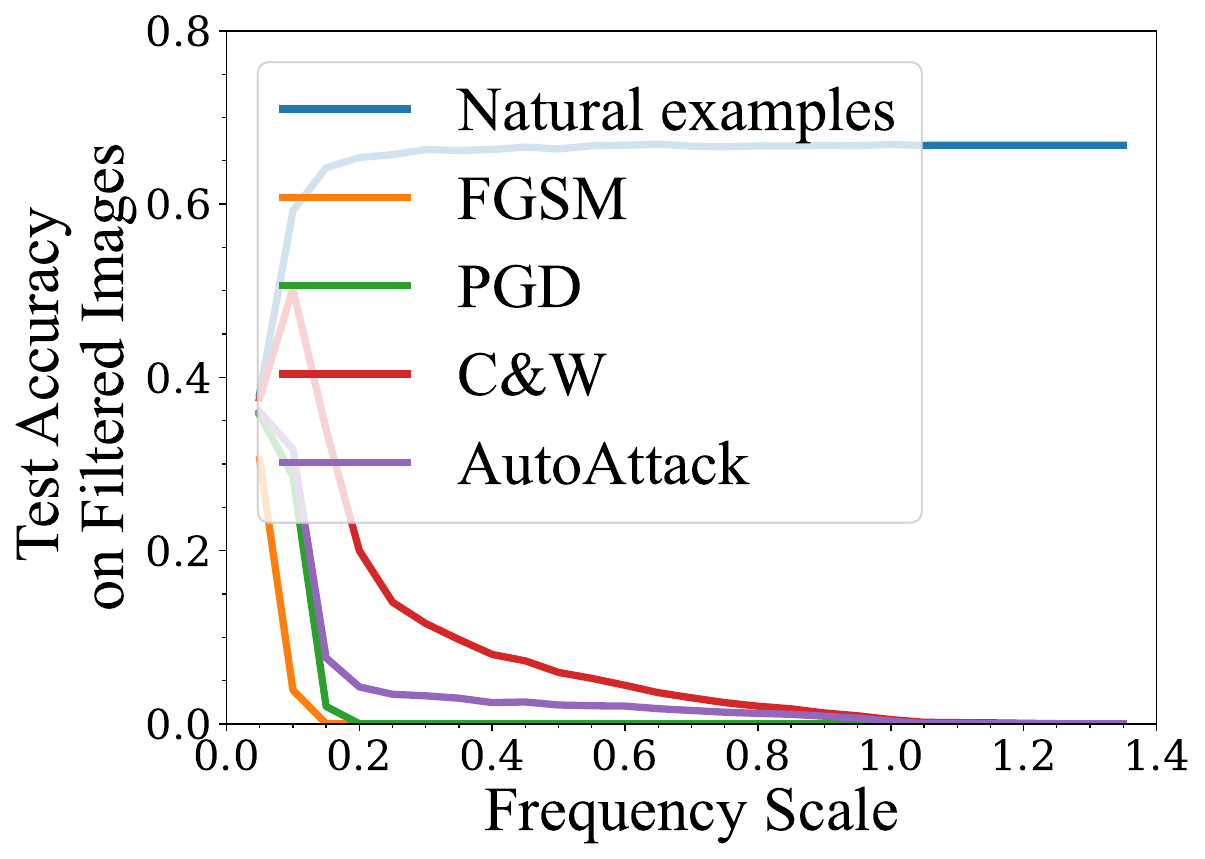}\label{fig:hfca}}
    \subfigure[Attack methods on CIFAR-100]{\includegraphics[width=0.24\linewidth]{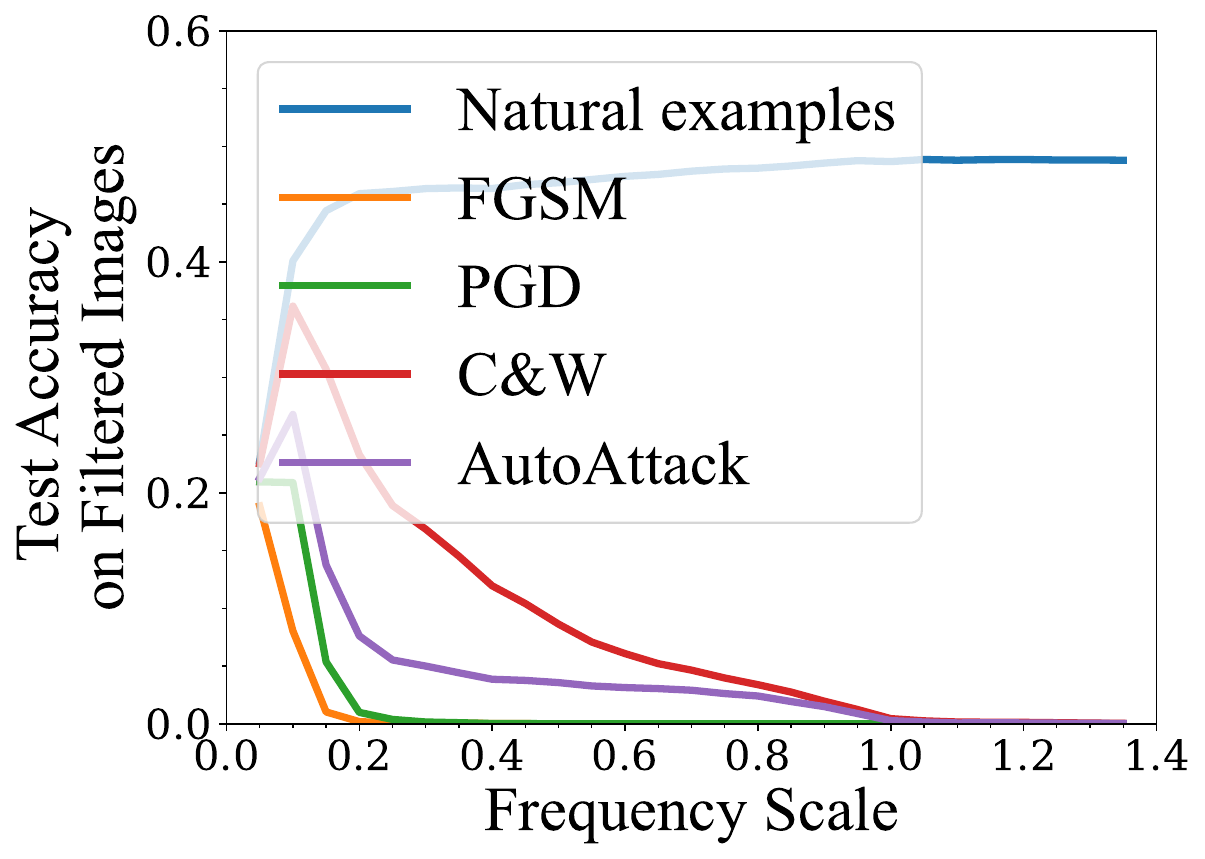}} 
    \subfigure[Attack methods on Tiny ImageNet]{\includegraphics[width=0.24\linewidth]{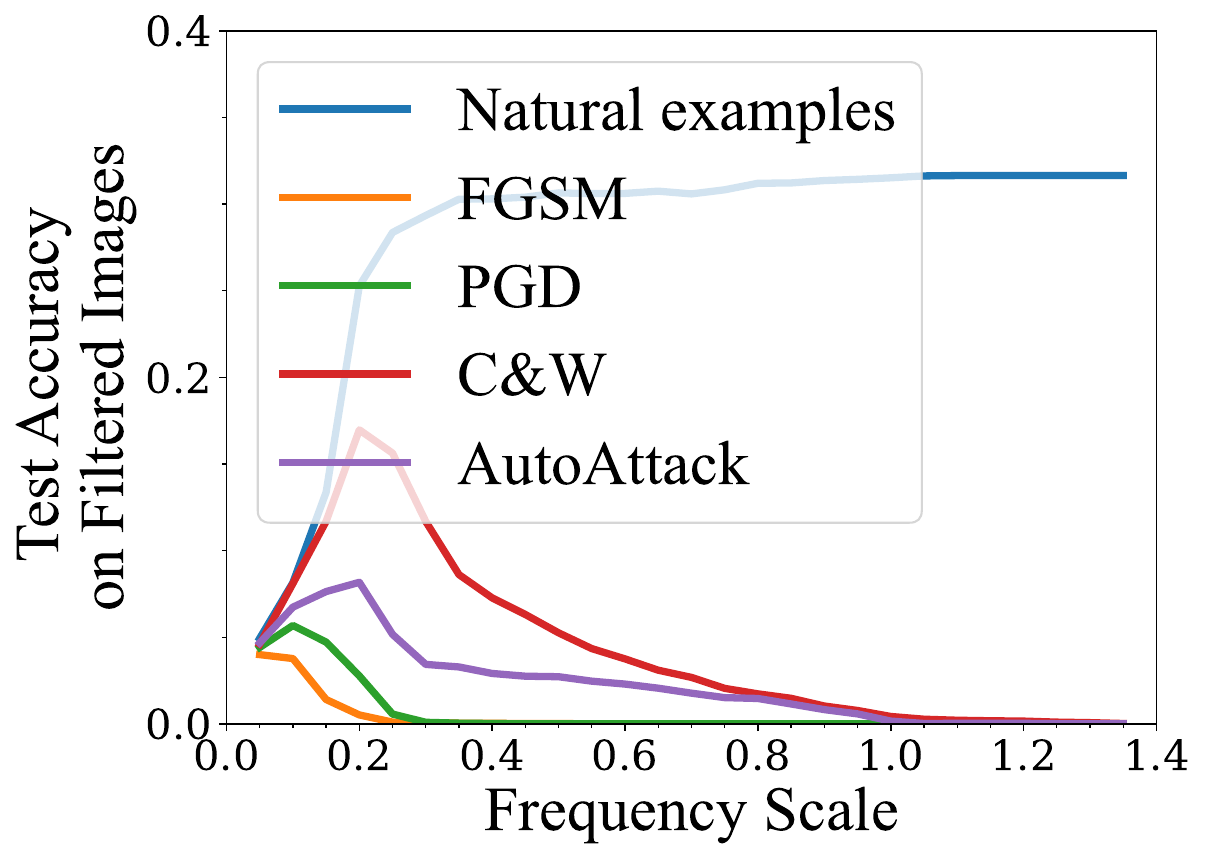}}\label{fig:hfcb} 
    \subfigure[Performance Gap]{\includegraphics[width=0.24\linewidth]{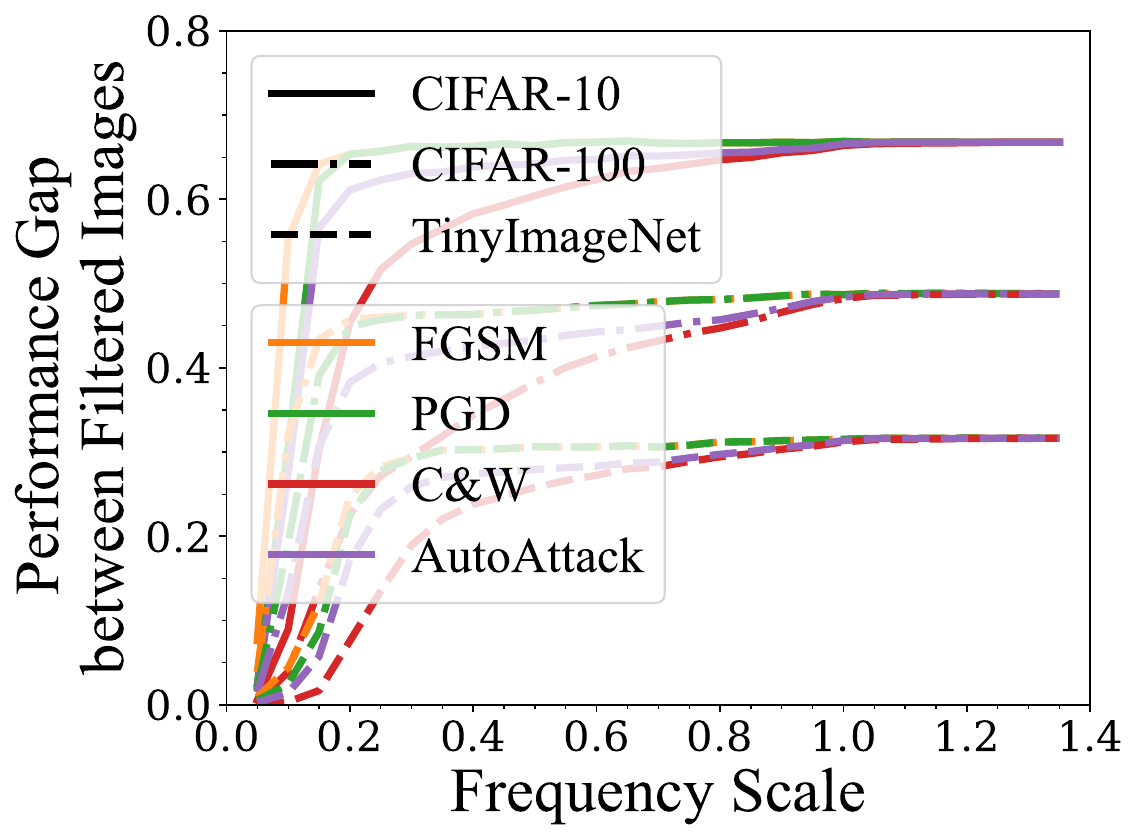}} 
    \vspace{-0.2cm}
    \caption{The differences of frequency components contribute to the performance gap between adversarial and natural examples on ViTs \textcolor{black}{across various attack methods}. We evaluated the accuracy of the filtered adversarial examples \textcolor{black}{generated using four different attack methods, the FGSM, C\&W, PGD, and AutoAttack methods, on multiple datasets}. (d) As higher frequency components are introduced, the performance gap between adversarial and natural examples increases \textcolor{black}{across these attack methods}.}
    \vspace{-0.2cm}
    \label{fig:ViTs_attacks}
\end{figure*}

\textcolor{black}{To validate the above observations, we conduct more experiments across multiple datasets and various attack methods. Fig.~\ref{fig:ConvNets_datasets} shows that the model performance \textit{w.r.t.} the filtered adversarial and natural examples on the CIFAR-100 and Tiny ImageNet datasets, where the adversarial examples are generated using the PGD method. Similarly, the performance gap between filtered adversarial/natural examples gradually
increases from zero to the gap between the model’s generalization and robustness. It is worth noting that due to variations in image size and resolution across datasets, the frequency range at which filtered adversarial examples achieve their peak performance may differ across different datasets. Fig.~\ref{fig:ConvNets_attacks} presents that the model performance \textit{w.r.t.} the filtered adversarial examples generated using the FGSM, C\&W, PGD, and AutoAttack methods on the ResNet-50~\cite{He16}. Among these methods, FGSM and PGD are gradient-based attacks, AutoAttack is an ensemble attack that includes PGD, and C\&W is an optimization-based attack. The results confirm that a similar phenomenon occurs across different attack methods. }

\begin{mdframed}[backgroundcolor=white!10,rightline=true,leftline=true,topline=true,bottomline=true,roundcorner=2mm,everyline=true,nobreak=false]  
\textbf{Proposal 1}: For Convolutional Neural Networks, the differences in mid- and high-frequency components between adversarial and natural examples play a critical role in their performance gap on models. These frequency components of adversarial examples exhibit their attack capabilities on models, and simply filter out these frequency components can effectively alleviate the vulnerability of models. This indicates that the researchers can focus more on the mid- and high-frequency components of the adversarial examples for detection and defence.
\end{mdframed}
\vspace{-4mm}


\subsection{Frequency Components of images on Transformers for \textbf{Q2}}

\begin{figure*}[htp]
    \centering
      \subfigure[ResNet-18 (ADV) ]{\includegraphics[width=0.24\linewidth]{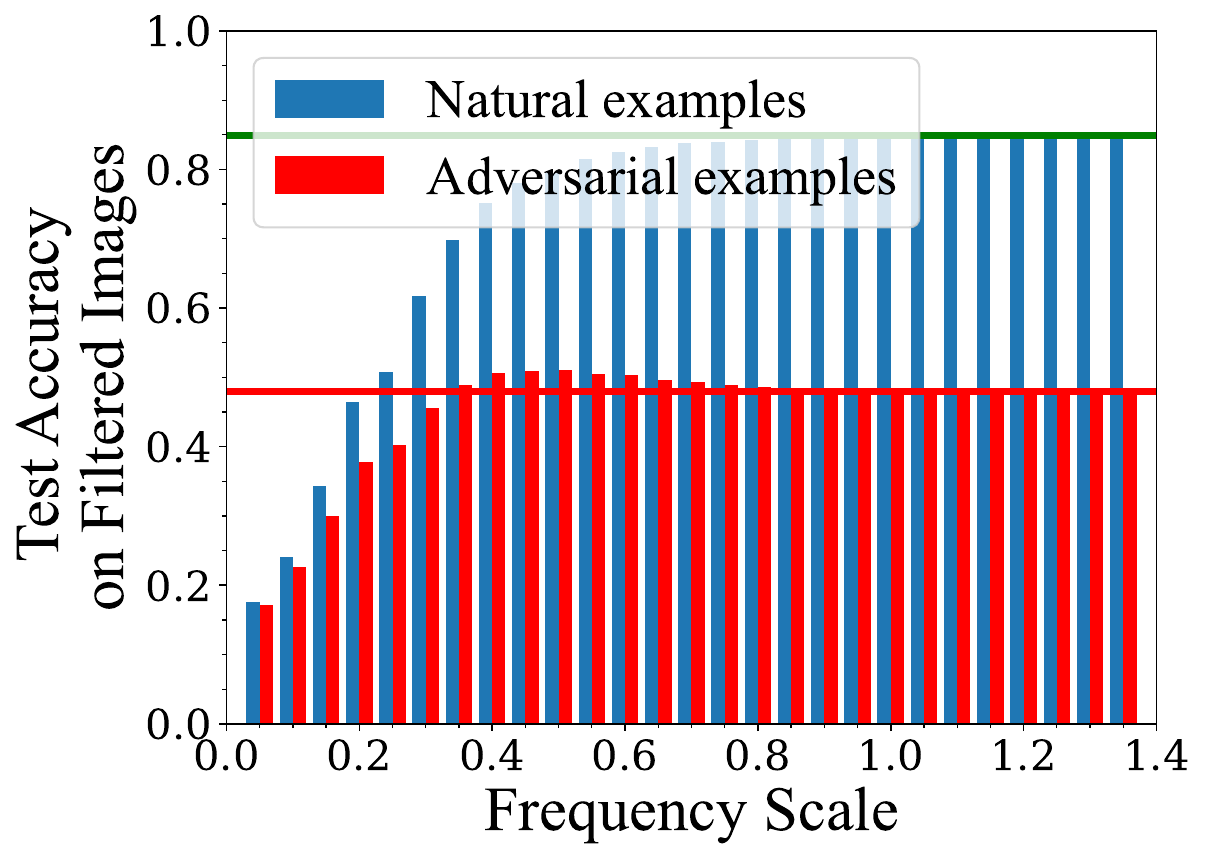}} 
    \subfigure[Wide ResNet-28-10 (ADV) ]{\includegraphics[width=0.24\linewidth]{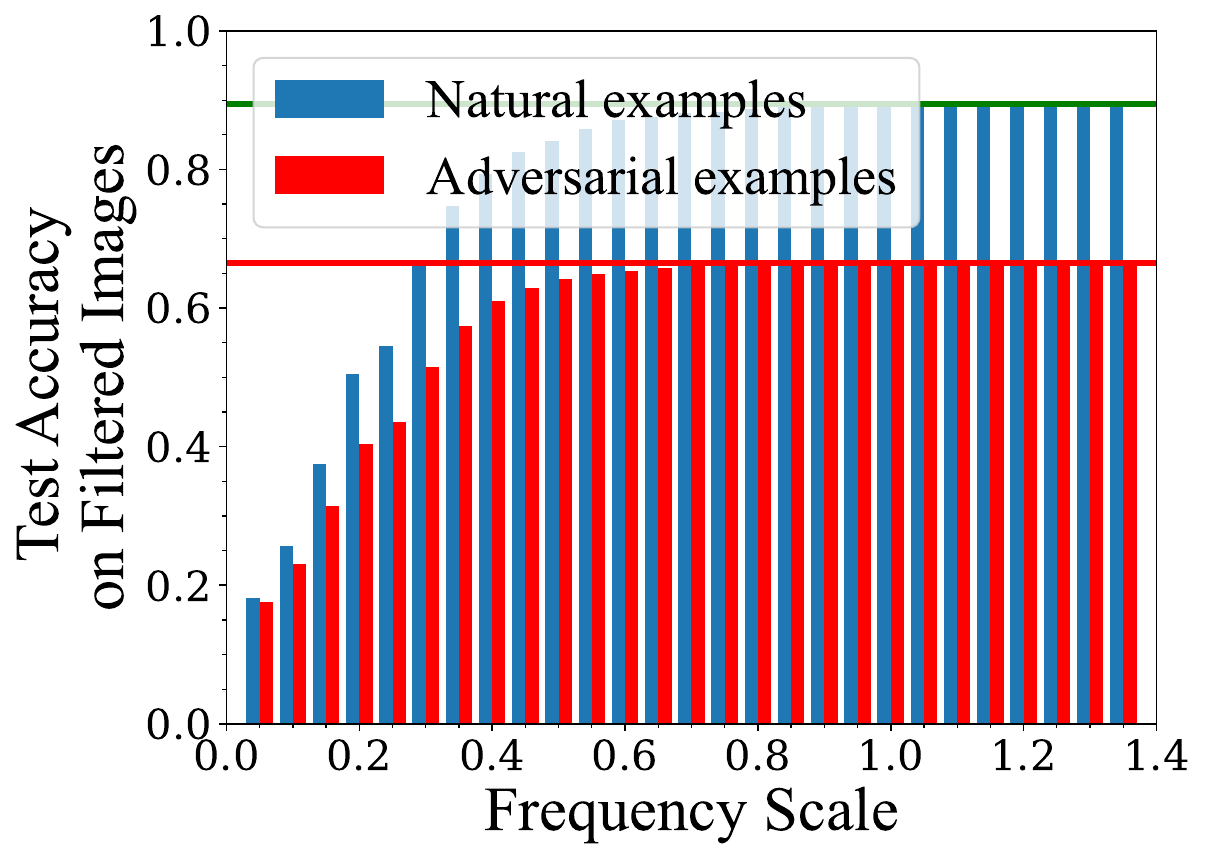}} 
    \subfigure[Merged Images (STD) ]{\includegraphics[width=0.24\linewidth]{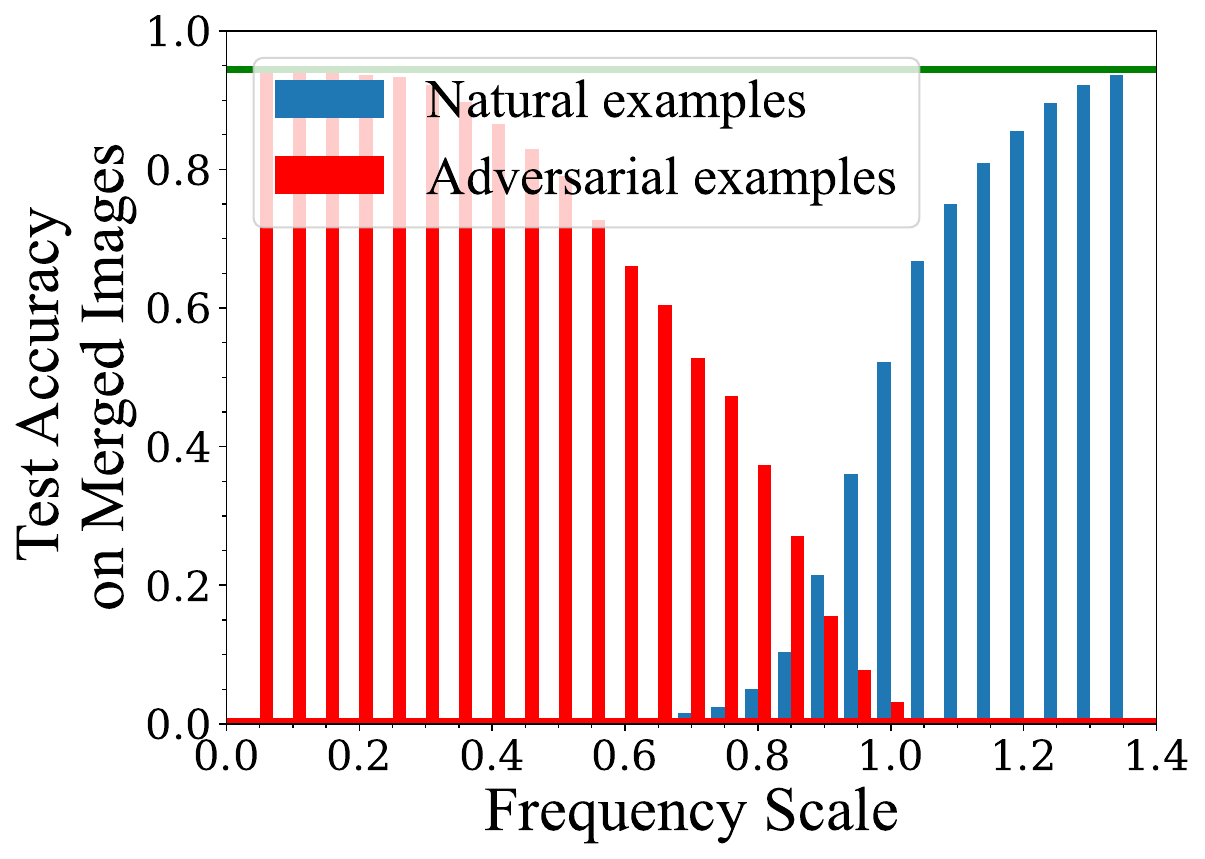}} 
    \subfigure[Merged Images (ADV) ]{\includegraphics[width=0.24\linewidth]{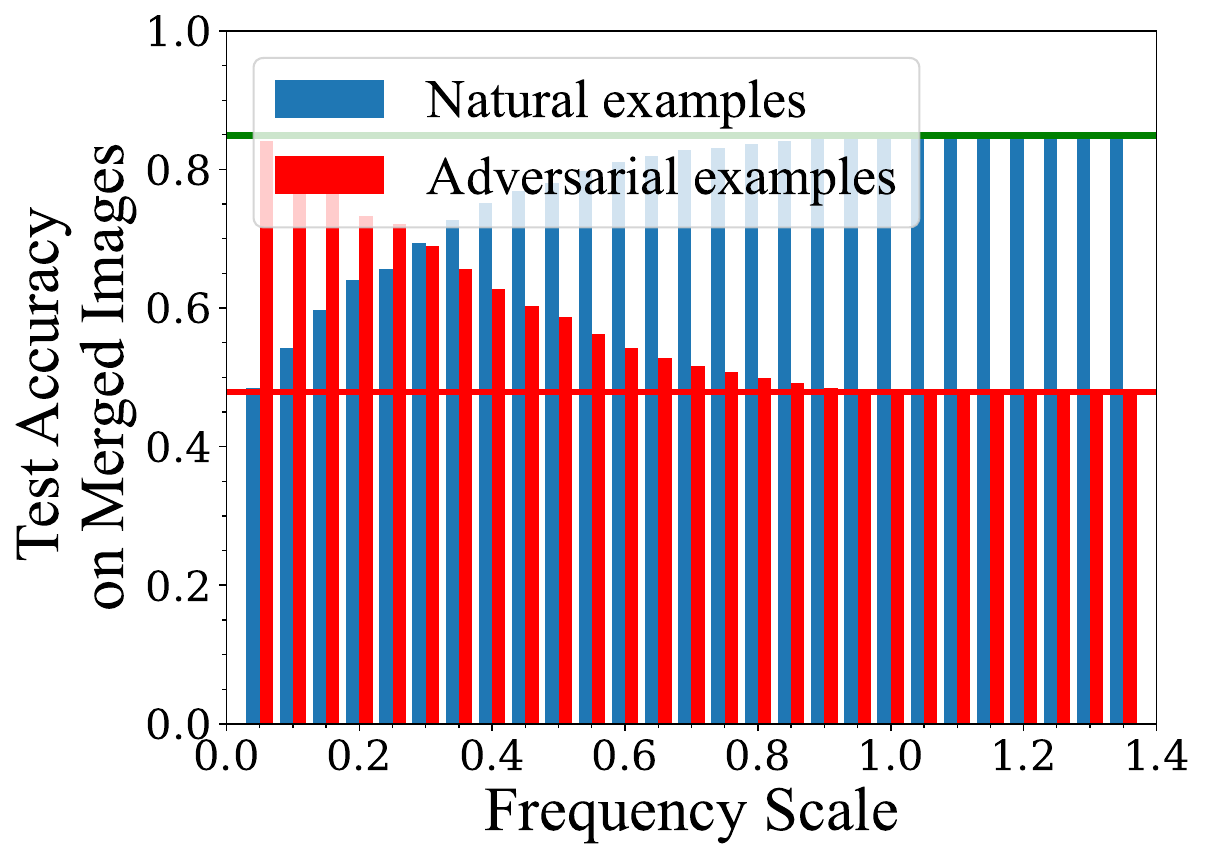}} 
    \vspace{-0.2cm}
    \caption{The differences in frequency components statistically contribute to the performance gap between adversarial and natural examples on adversarially-trained models. We tested the accuracy of the filtered adversarial/natural images on (a)-(b) different adversarially-trained (ADV) models. We tested the accuracy of the merged adversarial/natural images on both (c) the standard (STD) model and (d) the adversarially-trained (ADV) model.}
    \label{fig:ADV_models}
    \vspace{-0.15cm}
\end{figure*}

\begin{figure*}[!htbp]
    \centering
    \vspace{-0.2cm}
    \subfigure[ images]{\label{fig:fda}\includegraphics[width=0.15\linewidth]{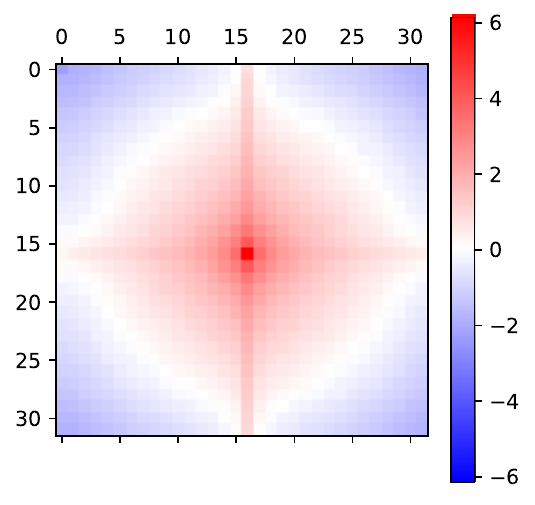}}   
    \subfigure[adv. (STD)]{\includegraphics[width=0.15\linewidth]{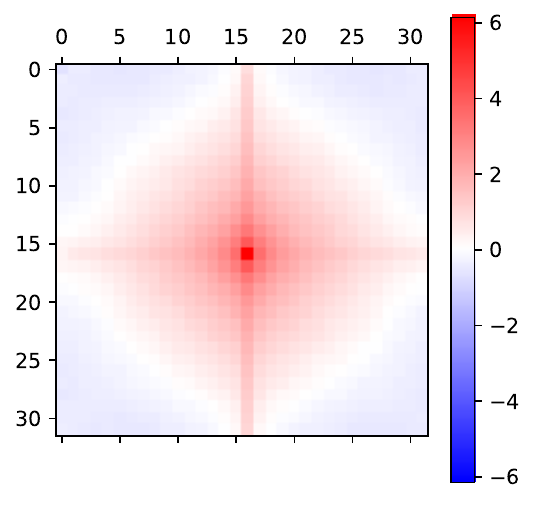}\label{fig:fdb}} 
    \subfigure[diff.  (STD)]{\includegraphics[width=0.157\linewidth]{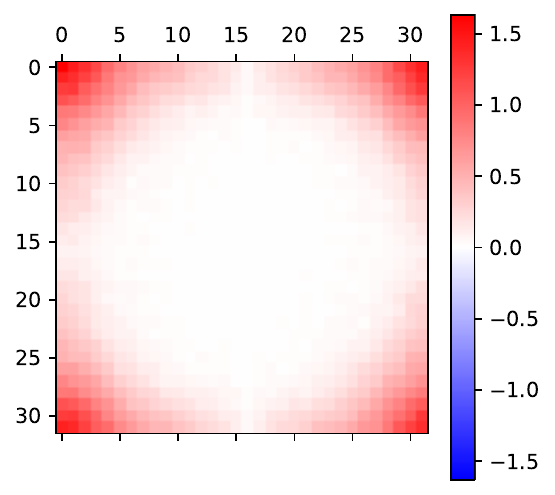}\label{fig:fdc}} 
    \subfigure[adv.  (ADV)]{\includegraphics[width=0.15\linewidth]{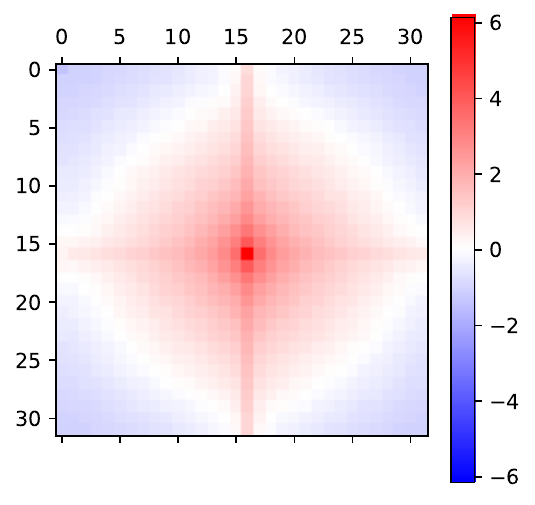}\label{fig:fdd}} 
    \subfigure[diff.  (ADV)]{\label{fig:fde}\includegraphics[width=0.157\linewidth]{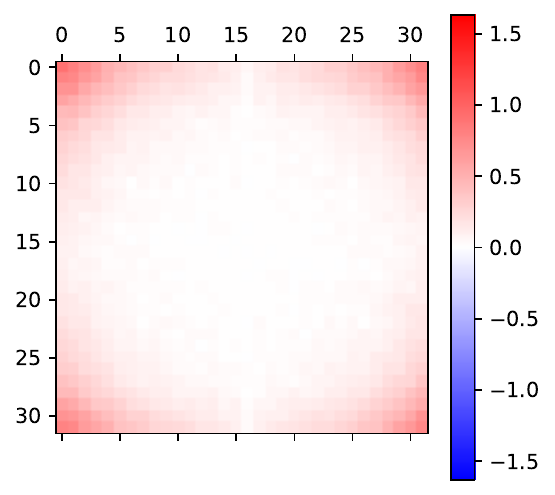}} 
    \vspace{-0.2cm}
    \caption{The average logarithmic amplitude spectrum of (a) 1000  three-channel images and (b) their adversarial examples generated by the standard (STD) model, where the corners represent high-freq components, and the colorbars represent the logarithmic amplitude $log(|\cdot|)$ (the redder the larger). (c) denotes the difference between (b) and (a), $log(|adv|)-log(|nat|)=log(|adv|/|nat|)$, where the write color of (c) represents equivalent, and the red represents $|adv|>|nat|$. (d) and (e) are on the adversarially-trained (ADV) model, and (e) denotes the difference between (d) and (a).
    }
    \label{fig:frequencydomain}
    \vspace{-0.3cm}
\end{figure*}

Fig.~\ref{fig:Transformers} shows that the model performance \textit{w.r.t.} the filtered adversarial/natural examples on various \textcolor{black}{Vision Transformers}, including ViT-B/16~\cite{dosovitskiy2020image}, Swin-B~\cite{liu2021swin}, and DeiT-B~\cite{touvron2021training}, on the CIFAR-10 dataset~\cite{Krizhevsky09}. \textcolor{black}{The adversarial examples are generated using the PGD method.} Similar as the phenomenon for ConvNets, as higher frequency components are introduced, the performance of filtered adversarial examples initially increases and subsequently declines to 0.0\%  (red line). Compared to ConvNets, the performance of filtered adversarial examples on Transformers declines even at low and mid frequencies, indicating that these frequency ranges contribute to the attack utility against Transformers. Fig.~\ref{fig:Transformers}(d) further illustrates that, statistically, the low- and mid-frequency components of both adversarial and natural examples have a significant differential impact on the model performance. As higher frequency components are incorporated, the performance gap between filtered adversarial and natural examples rapidly expands from 10-20\% to the gap between model generalization and robustness. 

\textcolor{black}{We further validate the above observations across more datasets and various attack methods on ViTs. Fig.~\ref{fig:ViTs_datasets} shows that the model performance \textit{w.r.t.} the filtered adversarial and natural examples on the CIFAR-100 and Tiny ImageNet datasets. Consistent with above findings, the low- and mid-frequency components contribute to the gap between the model’s generalization and robustness across different datasets. Furthermore, Fig.~\ref{fig:ViTs_attacks} presents the model performance \textit{w.r.t.} the filtered adversarial examples generated using different attacks. For Vision Transformers, adversarial examples generated by these four widely adopted attack methods predominantly exhibit their attack capabilities in the low- to mid-frequency regions.}



\begin{mdframed}[backgroundcolor=white!10,rightline=true,leftline=true,topline=true,bottomline=true,roundcorner=2mm,everyline=true,nobreak=false] 
\textbf{Proposal 2}: For Vision Transformers, the differences in low- and mid-frequency components between adversarial and natural examples play a critical role in their performance gap, and these frequency components of adversarial examples exhibit their attack capabilities on models. The researchers can focus more on the low- and mid-frequency components of adversarial examples to alleviate the vulnerability of Transformers, and narrow the performance gap between adversarial/natural examples.
\end{mdframed}

\subsection{Frequency Components of images on robust models for \textbf{Q3}}

We evaluate the model performance \textit{w.r.t.} the filtered adversarial/natural examples on adversarially-trained ResNet-18 and Wide ResNet-28-10. Fig.~\ref{fig:ADV_models}(a)-(b) show that for robust models, as higher frequency components are introduced, the model performance of filtered adversarial examples finally reaches model robustness without a rapid drop. Compared to standard models in Fig.~\ref{fig:ConvNets}(a)-(b), the low- and mid- frequency components of adversarial and natural examples initially exhibit the difference in their impact on the model performance. As higher frequency components are incorporated, the performance gap between filtered adversarial examples and natural examples increases rapidly to reach the gap between model generalization and robustness. 

\noindent\textbf{Comparing the frequency components of adversarial examples on standard and robust models.} We further investigate the the frequency-swapped images between natural and adversarial examples on the CIFAR-10 dataset. Fig.~\ref{fig:ADV_models}(c)-(d) clearly shows the model performance \textit{w.r.t.} the merged adversarial/natural examples on the standard and adversarially-trained ResNet-18, respectively. 
As higher frequency components of adversarial examples are incorporated, the classification accuracy of the merged adversarial examples gradually decreases from model generalization (green line) to model robustness (red line). Similarly, as higher frequency components of natural examples are incorporated, the
classification accuracy of the merged natural examples increases from model robustness (red line) to model generalization (green line). Experiments on merged images clearly illustrate the mid- and high- frequency components of adversarial examples significantly corrupt models on standard models, while the low- and mid- frequency components of adversarial examples corrupt models on robust models.



\noindent\textbf{The statistical differences in frequency components between adversarial and natural examples.} Fig.~\ref{fig:frequencydomain} visualizes the logarithmic amplitude spectrum of adversarial and natural examples. It shows that the difference in the frequency domain between two types of examples is concentrated in their high-frequency region. As Fig.~\ref{fig:frequencydomain}(c) shows, compared to natural examples in Fig.~\ref{fig:frequencydomain}(a), the high-frequency components of adversarial examples that generated from standard models are more pronounced in Fig.~\ref{fig:frequencydomain}(b). 
Fig.~\ref{fig:fdd} and \ref{fig:fde} further show that high-frequency components of adversarial examples that generated from adversarially-trained models are less than those from standard models, yet still more than natural examples'. Besides, Fig.~\ref{fig:fde} shows that adversarial examples generated by adversarially trained models exhibit more low- and mid-frequency components.

\begin{mdframed}[backgroundcolor=white!10,rightline=true,leftline=true,topline=true,bottomline=true,roundcorner=2mm,everyline=true,nobreak=false] 
\textbf{Proposal 3}: For adversarially-trained models, low- and mid-frequency components contribute to the performance gap between adversarial and natural examples. To further enhance model robustness, more attention should be directed towards improving robustness against these low- and mid-frequency components in adversarial examples.
\end{mdframed}

\section{Conclusion}
In this study, we identify intriguing properties of adversarial examples from a frequency-domain perspective. We observe the following findings. (1) The performance gap of ConvNets and ViTs between adversarial and natural examples becomes increasingly pronounced as higher frequency components are introduced. (2) The model performance against filtered adversarial examples initially increases to a peak and subsequently decreases to model robustness. (3) For ConvNets, the differences in mid- and high-frequency components contribute to their performance gap on models, whereas for ViTs the low- and mid-frequency components of adversarial examples exhibit their attack capabilities on models. Therefore, to further enhance the robustness of adversarially-trained models, more attention should be paid to the low-and mid-frequency components of adversarial examples. 






\bibliographystyle{IEEEbib}
\bibliography{icme2025references}


\end{document}